\documentclass[acmlarge]{acmart}

\AtBeginDocument{%
  \providecommand\BibTeX{{%
    \normalfont B\kern-0.5em{\scshape i\kern-0.25em b}\kern-0.8em\TeX}}}

\setcopyright{acmlicensed} 
\copyrightyear{2025} 

\acmJournal{IMWUT} 
\acmYear{2025} 
\acmVolume{9} 
\acmNumber{1} 
\acmArticle{11} 
\acmMonth{3} 
\acmDOI{10.1145/3699732} 

\usepackage{amsmath,amsthm,amsfonts,bm}
\usepackage{algorithmic}
\usepackage{graphicx}
\usepackage{textcomp}
\usepackage{xcolor,color}
\usepackage{booktabs} 
\usepackage{tabularx}
\usepackage{float}
\usepackage[ruled,linesnumbered,titlenumbered]{algorithm2e}

\usepackage{makecell,multirow}

\usepackage{framed}
\usepackage{threeparttable}
\usepackage{caption}
\usepackage{subcaption}

\newtheorem{theorem}{Theorem}

\newcommand{\tool}{\textsc{LGL-BCI}}

\begin{document}

\title{{\tool}: A Motor-Imagery-Based Brain--Computer Interface with Geometric Learning}


\author{Jianchao Lu}
\authornote{Both authors contributed equally to this research.}
\email{jianchao.lu@mq.edu.au}
\orcid{0000-0003-0788-1448}
\author{Yuzhe Tian}
\authornotemark[1]
\email{yuzhe.tian@hdr.mq.edu.au}
\orcid{0000-0002-5742-7414}
\affiliation{%
  \institution{Macquarie University}
  \department{School of Computing}
  \city{Macquarie Park}
  \state{NSW}
  \country{Australia}
  \postcode{2109}
}

\author{Yang Zhang}
\email{yang.zhang@mq.edu.au}
\email{yang.zhang@unt.edu}
\orcid{0000-0001-6821-2710}
\affiliation{%
  \institution{Macquarie University}
  \department{School of Computing}
  \city{Macquarie Park}
  \state{NSW}
  \country{Australia}
  \postcode{2109}
}
\affiliation{%
  \institution{University of North Texas}
  \department{Department of Information Science}
  \city{Denton}
  \state{Texas}
  \country{USA}
  \postcode{76207}
}

\author{Quan Z. Sheng}
\email{michael.sheng.mq.edu.au}
\orcid{0000-0002-3326-4147}
\affiliation{%
  \institution{Macquarie University}
  \department{School of Computing}
  \city{Macquarie Park}
  \state{NSW}
  \country{Australia}
  \postcode{2109}
}

\author{Xi Zheng}
\authornote{Corresponding author.}
\email{james.zheng@mq.edu.au}
\orcid{0000-0002-2572-2355}
\affiliation{%
  \institution{Macquarie University}
  \department{School of Computing}
  \city{Macquarie Park}
  \state{NSW}
  \country{Australia}
  \postcode{2109}
}

\begin{abstract}
Brain--computer interfaces are groundbreaking technology 
whereby brain signals are used to control external devices.
Despite some advances in recent years, 
electroencephalogram (EEG)-based motor-imagery tasks face challenges, 
such as amplitude and phase variability and complex spatial correlations,
with a need for smaller models and faster inference.
In this study, we develop a prototype, 
called the Lightweight Geometric Learning Brain--Computer Interface ({\tool}), 
which uses our customized geometric deep learning architecture for swift model inference without sacrificing accuracy.
{\tool} contains an EEG channel selection module via a feature decomposition algorithm 
to reduce the dimensionality of a symmetric positive definite matrix, 
providing adaptiveness among the continuously changing EEG signal.
Meanwhile, a built-in lossless transformation helps boost the inference speed.
The performance of our solution was evaluated using two real-world EEG devices and two public EEG datasets.
{\tool} demonstrated significant improvements,
achieving an accuracy of 82.54\% compared to 62.22\% for the state-of-the-art approach.
Furthermore, {\tool} uses fewer parameters ($64.9K$ vs. $183.7K$),
highlighting its computational efficiency.
These findings underscore both the superior accuracy and computational efficiency of {\tool},
demonstrating the feasibility and robustness of 
geometric deep learning in motor-imagery brain--computer interface applications.
\end{abstract}




\maketitle

\section{Introduction}
\label{sec:introduction}

Recent growth in electroencephalogram (EEG)-based motor imagery (MI) research,
fueled by its transformative potential, has notably advanced brain--computer interfaces (BCIs).
BCIs provide a unique interface for individuals to control external devices using brain signals~\cite{yuan2014brain},
significantly benefiting those with motor disabilities~\cite{palumbo2021motor}.
For example, paralyzed patients can use MI-BCIs to
operate robotic limbs, communicate, and control computers.

Several mainstream solutions are currently employed. For instance, \citet{koles1990spatial}
leveraged common spatial patterns
(CSPs) to improve the signal-to-noise ratio (SNR) of EEG signals
and amplify oscillatory brain activities for pre-feature extraction. \citet{bang2021spatio}
employed
convolutional neural networks (CNNs)
to capture the spatiotemporal frequency features of neural signals and optimize EEG data translation.
Despite rapid advancements in deep learning frameworks for EEG signal processing,
MI-BCIs face enduring challenges.

\begin{enumerate}
\item[] \textbf{Challenge 1. Variability in amplitude \& phase.}
EEG signals are non-stationarity by nature, making them difficult to extract consistently.
Individuals can produce varying EEG signals under similar conditions,
known as inter-subject variability.
Even within a single individual,
factors like fatigue or attention shifts can cause variations,
known as intra-subject variability~\cite{roy2022multi}.

\item[] \textbf{Challenge 2. Complex spatial correlation.}
The volume conduction effect causes signals
from a specific brain area to be detected by multiple electrodes~\cite{anastasiadou2019graph,anzolin2018effect}.
It is necessary to mitigate this effect for accurate EEG-based brain network analysis.
Moreover, EEG data, being multi-dimensional from numerous scalp electrodes~\cite{mai2021affective,lu2022pearnet},
require a non-Euclidean techniques to retrieve spatial features relevant to MI-BCI,
as opposed to traditional Euclidean-based methods~\cite{ju2022tensor,congedo2017riemannian}.

\item[] \textbf{Challenge 3. Model size \& inference speed.}
Effective MI-BCIs demand real-time operation,
but large, complex deep-learning models require power-draining computing hardware,
leading to non-real-time responses, 
due to the latency caused by computation.
This renders them unsuitable for real-time usage, 
especially in computationally constrained BCI applications,
for example with wearable devices for long-term BCI usage.
Thus, it is essential to minimize model size and inference speed.
\end{enumerate}

While current methods suffer from model performance variability
and high computational complexity that limits mobile deployment,
we unveil the Lightweight Geometric Learning Brain--Computer Interface ({\tool}),
a prototype leveraging geometric deep learning and optimizing for MI tasks on resource-constrained devices.
Our approach also employs non-Euclidean metric spaces like the 
symmetric positive-definite (SPD) manifold~\cite{xie2016motor,corsi2022functional,lotte2018review}
for enhanced EEG representation. 
The proposed solution involves four modules:
(1) SPD manifold construction,
(2) spatial feature extraction within it,
(3) key EEG channel selection through a novel dimension-reduction technique using geometry-aware mapping to tangent spaces,
and
(4) employing a CNN to extract temporal features from these spaces.
Our main contributions are as follows:

\begin{itemize}

\item Pioneering Geometric Learning in EEG:
To the best of our knowledge, we are the first to leverage geometric deep learning for real-world EEG signal processing with consumer-grade devices. 
This novel approach utilizes the SPD manifold, 
which provides a geometrically principled way to handle the spatial correlations in EEG data, 
capturing the complex, non-linear dynamics of brain signals.

\item Addressing Complexity and Efficiency:
We propose a novel approach to reduce the dimensionality of the SPD matrix in EEG analysis 
by deriving a function that links geometric and Euclidean distance matrices. 
Using eigenvalue decomposition, we extract the top $d$ eigenvectors to retain the most significant EEG channels. 
This is one of the first low-rank approximation techniques applied to geometric deep learning on SPD manifolds, and it has proven to be efficient.

\item Innovative Multi-Head Bilinear Transformation:
We apply multi-head bilinear transformation in the tangent space of the SPD manifold, 
rather than the traditional multi-head approach based on Euclidean space. 
This allows us to capture complex non-linear relationships within the SPD manifold, 
marking one of the first attempts to do so.

\item Extensive Evaluations:
Compared to existing approaches primarily tested on public datasets, 
our approach was validated through comprehensive experimental studies, 
including evaluations on two public datasets and real-world scenarios with consumer-grade hardware. 
The results are promising, showing competitive accuracy ($72.51\%$ vs. $72.56\%$ for MI-KU and $76.95\%$ vs. $77.55\%$ for BCIC-IV-2a) with fewer parameters. 
In real-world scenarios, our tool exhibits significant improvements, 
achieving higher accuracy ($82.54\%$ vs. $62.22\%$ of the state-of-the-art approach) with fewer parameters ($64.9K$ vs. $183.7K$) and reduced inference time (up to 4 ms faster).

\end{itemize}

In the remainder of this paper, 
we first provide background knowledge and motivations in Section~\ref{sec:motivation}.
We present our {\tool} approach in Section~\ref{sec:approach}.
Following that, we explore the experimental design in Section~\ref{sec:exp}
and present a comprehensive discussion of the experimental results in Section~\ref{sec:result}.
In Section~\ref{sec:future}, 
we discuss the limitations of {\tool} and briefly point out future research directions.
Section~\ref{sec:conclusion} concludes.

\section{Motivations}
\label{sec:motivation}

\subsection{Empowering Disabilities with EEG Real-World Deployment}

\subsubsection{EEG-Controlled Mobility Assistance Devices}
\label{sec:eeg-wheelchair}
Conventional mobility assistance options, such as canes, rollators, and manual or powered wheelchairs,
require the users to retain some basic motor functions for control~\cite{kapsalis2024disabled}.
Meanwhile, some diseases can lead to irreversible motor function loss,
while some accidents could also lead to a high loss of motor function, even complete paralysis~\cite{fortuna2017healing}.
It would be a huge challenge for these users to leverage the conventional options.

As we set wheelchairs as the moving target, at least four commands for controlling are needed: Forward, Stop, Left Rotation, and Right Rotation. These align with the basic movements
for conventional wheelchair usage ~\cite{togni2022turning,bickelhaupt2018effect,bertolaccini2018influence}.
However, the direct thoughts of these movements (the thoughts of Forward, Stop, Left Rotation, and Right Rotation) are
generated in a deep layer of the brain~\cite{liao2014decoding}.
At the current stage, they are mostly carried out through electrocorticography (ECoG),
which requires surgical brain implants~\cite{lal2004methods,schalk2007decoding}.
As a non-invasive method that is more feasible, EEG signals can be gathered from the scalp.
Yet the skull interferes with signals from deep layers of the brain~\cite{ebrahiminia2022multivariate},
and this makes it hard to extract these thoughts through EEG signals.

On the other hand, previous research has found that
an EEG-based MI-BCI is promising for several classes of movement~\cite{wang2023ifnet,ma2023temporal,mane2021fbcnet}.
Thus, we built the controlling scheme with a motor-imagery encoding method.
We chose daily activities that commonly appear
in real life to build up a real-world applicable prototype.
Meanwhile, we ensured a sufficient number of imagery movements
for encoding the four basic movements for controlling the wheelchair.

The visually inspired EEG signals collected in our experiment
enable us to assign not only directional encoding for these
visual signals (e.g. Left Rotation and Right Rotation),
but also specific encodings that can be received by remote devices to control various home appliances.
This includes turning TVs or heaters on and off and making emergency calls.
Thus, evaluating such visual motor imagery tasks is crucial for customized encoding,
not only for directional commands of wheelchairs
but also for controlling various remote devices for disabled individuals.
The details are given in Section~\ref{sec:exp}.

\subsubsection{Problems with Current EEG Devices}

Current research on EEG-controlled wheelchairs requires participants
to undergo an adequate procedure that tailors the performance for better EEG command decoding performance~\cite{tonin2022learning}.
Lab-grade equipment and devices are costly
and require maintenance and precise adjustments. 
As such, they are not feasible for the users.
On the other hand, current consumer-grade EEG devices do not have
standardized signal outputs. Their output range varies from brand to brand.
As a result, a less generalized framework would suffer from these variations.
Therefore, we propose our {\tool} prototype,
which leverages EEG signal outputs from different brands of devices.
The details are provided in Section~\ref{sec:hardware}.

\subsection{Geometric Deep Learning and Current Implementation in MI-BCIs}

Conventionally, by introducing various spatial filtering methods,
the signal-to-noise ratio (SNR) can be improved to amplify
oscillatory brain activities for pre-feature
extraction~\cite{koles1990spatial, lu2010regularized, lotte2010regularizing}.
However, the effectiveness of these techniques is limited when dealing with features from complex and even unrelated neural oscillations \cite{lawhern2018eegnet}.
Meanwhile, traditional deep learning methods like CNNs
have proven effective at addressing the low SNR of EEG signals
in the BCI domain \cite{sakhavi2018learning,bang2021spatio,stieger2021benefits}.
But due to structural differences between images and EEG signals,
CNN feature expression in motor-imagery tasks is restricted by the non-Euclidean nature of
EEG signals \cite{bruna2013invariant,lecun2015deep,bronstein2017geometric}.
The main reason for this is that conventional deep learning methodologies
such as convolutional neural networks (CNNs) or recurrent neural networks (RNNs)
are inherently designed for grid-like data structures like images and sequential data stream.
However, in many real-world scenarios and scientific research areas,
the data that need to be processed are distributed in non-Euclidean space by nature: e.g. molecular structures and EEG signals.
These data are better presented in forms like graphs or manifolds.
Compared with conventional techniques,
geometric deep learning could help better extract the relationships
and patterns from non-Euclidean data~\cite{bronstein2017geometric}.

To handle non-Euclidean data structures (which are inherent in EEG signals),
geometric learning methods based on Riemannian geometry are widely adopted~\cite{jang2021riemannian,wang2015multi,singh2022data}.
If data points could be presented as SPD matrices,
the data would form a space known as the SPD manifold.
Within the non-Euclidean data,
Riemannian geometry helps elucidate curved spaces,
where the notion of straight lines (geodesics) is not the same as in Euclidean spaces~\cite{harandi2017dimensionality}.
Meanwhile, geodesics, which are the shortest paths between two points,
are the key to navigating these spaces.
In the SPD manifold, the tangent space at a specific point is crucial,
serving as a bridge between complex curvatures and simple linear operations.

In Riemannian geometry-based geometric deep learning,
some essential operations are also used in our work.
The first commonly used operation is BiMap,
which transforms spatial covariance matrices within each frequency passband
for the input EEG signal via \( W \cdot X^{ij} \cdot W^{T} \),
utilizing a full column-rank matrix \( W \).
Another commonly used operation is Riemannian batch normalization,
which is a variant of conventional batch normalization for data on Riemannian manifolds,
ensuring that the data stay within the manifold during training, improving and accelerating convergence.
The third operation is an activation function
in the form of a ReEig layer,
akin to ReLU in traditional deep learning,
rectifying input SPD matrices with a non-linear function applied to their elements or eigenvalues.
The fourth operation is a LogEig layer that maps elements of the SPD manifold to their tangent space for efficient processing, as shown in Eq.~\ref{eq:log-x}:
\begin{equation}
\log(X_{i}) = U \, \mathrm{diag}(\lambda_{1}, \ldots, \lambda_{d}) \, U^{T},
\label{eq:log-x}
\end{equation}
where \( i = 1,\ldots,d \), \( U \) is the orthogonal matrix of eigenvectors of \( X_i \),
and \( \lambda_1, \ldots, \lambda_d \) are the eigenvalues of \( X_i \),
while \(\mathrm{diag}\) is a function for creating a diagonal matrix.

Considering that EEG signals are non-Euclidean data,
leveraging geometric deep learning, especially Riemannian geometry and the SPD manifold,
could help simplify EEG data interpretation~\cite{lin2019riemannian}
and mitigate some of the challenges like the variability in amplitude and phase (Challenge 1)
and the spatial correlation complexity (Challenge 2).
Riemannian geometry-based geometric
learning methods have already been employed in MI-BCIs~\cite{huang2017riemannian,ju2022tensor}.
Riemannian geometry allows EEG data mapping onto a Riemannian manifold,
yielding metrics resistant to outliers and noise~\cite{pan2022matt}.
It also addresses the so-called swelling effect with the larger determinant of
Euclidean mean averaging from SPD matrices~\cite{yger2016riemannian}.
The manifold metrics exhibit invariance properties,
enhancing model generalization for complex EEG signals~\cite{congedo2017riemannian}.
Recent MI-BCIs~\cite{ju2022tensor, ju2022graph} primarily design
a feature extraction module based on SPD manifolds using Riemannian geometry,
where SPD matrices capture spatial correlations among EEG channels that are
crucial for understanding brain activity.
These matrices, as Riemannian manifolds,
possess specific invariance properties beneficial in practical applications,
such as invariance under invertible linear transformations,
to address amplitude and phase variability.

\subsection{Insights}

The direct application of Riemannian geometry-based geometric learning
for feature extraction from EEG signals is not applicable to MI-BCI applications
with an increasing number of channels of input signals \(n\),
due to the escalated storage \(O(n^{2})\) and computational demands \(O(n^{3})\)~\cite{huang2017riemannian}.

To reduce the model size and improve the inference speed,
we propose the Lightweight Geometric Learning Brain--Computer Interface ({\tool}).
In a nutshell, {\tool} completes the two
challenging tasks by leveraging eigenvalue decomposition and
manifold--Euclidean space mapping.
By leveraging matrix eigenvalue decomposition,
the SPD matrix dimension can be reduced while retaining the original information~\cite{zhang2023evaluation}.
{\tool} preserves inherent distance relationships
through a geometry-aware transform while trimming the input size.
It also employs unsupervised multi-bilinear transformation,
bypassing manual labeling and aiding temporal feature extraction on the tangent space, thereby
enhancing computational efficiency without major information loss.
Meanwhile, as inferences in Euclidean space are faster than in the manifold,
by mapping the relationship from the manifold to Euclidean space,
we perform a lossless transformation to help map from the manifold to Euclidean space. 
A detailed proof is provided in Section~\ref{sec:approach}.
{\tool} not only tackles the low SNR issue but also
explores the non-Euclidean relationships between EEG channels.
Our prototype offers the possibility of a lightweight interface
that considers both inference speed and network size.

\section{Related Work}
\label{sec:related}

\subsection{Geometric Deep Learning in MI-BCIs}

Data from
widely deployed real-world sensors
are naturally distributed in non-Euclidean spaces
and are often
represented as graphs or manifolds,
e.g. EEG signals.
Compared with other methodologies,
geometric deep learning could better extract the relationships
and patterns with non-Euclidean data~\cite{bronstein2017geometric}.
To handle non-Euclidean data structures effectively, geometric learning methods,
mainly based on Riemannian geometry, are widely
adopted~\cite{congedo2017riemannian,huang2017riemannian, yger2016riemannian, jang2021riemannian}.
Riemannian geometry helps to explain curved spaces,
notably the SPD manifold.
Meanwhile, geodesics, which are the shortest paths between any two points,
are the key to navigating these spaces.
In the SPD manifold, the tangent space at a specific point is crucial,
serving as a bridge between complex curvatures and simple linear operations.

As EEG signals are non-Euclidean data, leveraging geometric deep learning and multi-channel data analysis is essential.
Plenty of Riemannian geometry-based geometric
learning methods have already been employed in MI-BCI~\cite{huang2017riemannian,ju2022tensor}.
Riemannian geometry allows EEG data mapping onto a Riemannian manifold,
yielding metrics resistant to outliers and noise~\cite{pan2022matt}.
Riemannian geometry also mitigates the so-called swelling effect,
which occurs when averaging SPD matrices in Euclidean space. This results in a mean matrix with an artificially large determinant.
Riemannian geometry provides a more accurate method of averaging within
the curved space of SPD matrices~\cite{yger2016riemannian}.
The manifold metrics exhibit invariance properties,
enhancing model generalization for complex EEG signals~\cite{congedo2017riemannian}.
Recent MI-BCIs~\cite{ju2022tensor, ju2022graph} typically design
the feature extraction module based on SPD manifolds using Riemannian geometry,
where SPD matrices capture spatial correlations among EEG channels that are crucial for understanding brain activity.
These matrices, as Riemannian manifolds,
possess specific invariance properties beneficial in practical applications,
such as invariance under invertible linear transformations. These advantages are
useful for addressing amplitude and phase variability.

Our proposed work leverages geometric deep learning for
EEG signal processing by utilizing the SPD manifold
to manage spatial correlations in EEG data.
Unlike existing approaches that often overlook computational complexity,
our work introduces an efficient channel selection module
and a multi-head bilinear transformation strategy,
ensuring high performance in resource-constrained environments.
This approach results in a robust and efficient solution for real-world EEG applications.

\subsection{Other Methods in MI-BCIs}

One of the key hurdles to effectively using BCIs
is the need for a substantial amount of labeled data to
fine-tune the classification processes.
Also, a good signal-to-noise ratio (SNR) is essential for
collecting and analyzing EEG data efficiently.
To fully utilize EEG activity associated with imagined movements,
spatial filtering proves to be invaluable, as highlighted by~\cite{mammone2023autoencoder}.
Among these techniques,
the common spatial pattern (CSP) approach, detailed by~\cite{ang2012filter},
is especially popular for EEG classification in MI-BCIs.
Recent research has shown that merging CSP-based spatial filtering algorithms
with transfer learning techniques greatly improves the performance of MI-BCIs
and shortens their calibration time,
as noted by~\cite{wei2023intra}.
Furthermore,~\citet{jiang2020temporal} introduced
methods for optimizing the timing of visual cues in EEG signal processing using CSP,
significantly enhancing CSP's effectiveness.
In addition,~\citet{dong2023scatter} developed new CSP variants like scaCSP to
address challenges such as noise sensitivity, non-stationarity,
and restrictions in binary classification.

Another effective strategy for reducing the SNR in EEG signals
within the BCI field is the use of convolutional neural networks (CNNs)
for EEG-based MI classification, as explored by~\cite{wang2024mi}.
Specifically, MI-BCIs focus on detecting and interpreting brain motor imagery patterns.
CNNs excel at capturing the spatiotemporal frequency features of neural signals,
thereby enhancing the interpretation of EEG data; see \cite{sakhavi2018learning,bang2021spatio,stieger2021benefits}.
More recent developments include CNN-based end-to-end deep learning models,
like the dual-domain CNN-based subject-transfer model by~\cite{jeong2022cnn}
and the multi-domain CNN model by~\cite{jeong2021multi},
which excel at extracting and classifying features from low-SNR EEG signals.
The low SNR in MI EEG signals poses a challenge when decoding movement intentions,
but this can be addressed using multi-branch CNN modules
that learn spectral-temporal domain features, as suggested by~\cite{jia2023model}.
However, the structural differences between images
and EEG signals limit a CNN's feature expression
in motor-imagery tasks due to the non-Euclidean nature of
EEG signals; see~\cite{bruna2013invariant,lecun2015deep,bronstein2017geometric}.

Our research primarily utilizes non-invasive EEG extraction,
which contrasts significantly with methods like electrocorticography (ECoG) that
require surgical interventions.
According to \citet{metzger2023high}, ECoG involves placing electrodes directly on the cortex,
yielding high-resolution signals suitable for detailed command decoding.
While ECoG provides enhanced signal clarity and decoding accuracy,
its invasive nature raises concerns regarding surgical risks and
long-term viability for everyday use.
By contrast, our EEG-based approach, while slightly less accurate,
offers a non-invasive alternative that is more adaptable to daily applications
without the inherent risks of surgery.

Further, our approach contrasts with steady-state visual evoked potential (SSVEP)-based BCI systems,
which require external visual stimuli to elicit brain responses.
\citet{chen2022electric} discuss the use of SSVEP in BCIs,
highlighting its efficacy in controlled environments.
However, the dependency on external stimuli can limit the application of SSVEP
in everyday settings where such stimuli might not be consistently present.
Our motor-imagery-based EEG system provides a user-driven approach,
relying solely on the individual's internal cognitive functions,
thereby enhancing the system's portability and usability in a variety of environments.

In summary, the practical applications of our findings are particularly relevant
for individuals with mobility impairments.
Unlike the aforementioned techniques, our motor-imagery-based EEG system allows users,
such as those with limited mobility in their limbs due to injuries or paralysis,
to control devices like wheelchairs using imagined movements.
This method translates into a significant enhancement in autonomy and quality of life,
offering a feasible BCI application outside of clinical settings.

\section{The {\tool} Approach} 
\label{sec:approach}

\begin{figure*}[!tb]
\centering
\includegraphics[width=0.95\textwidth]{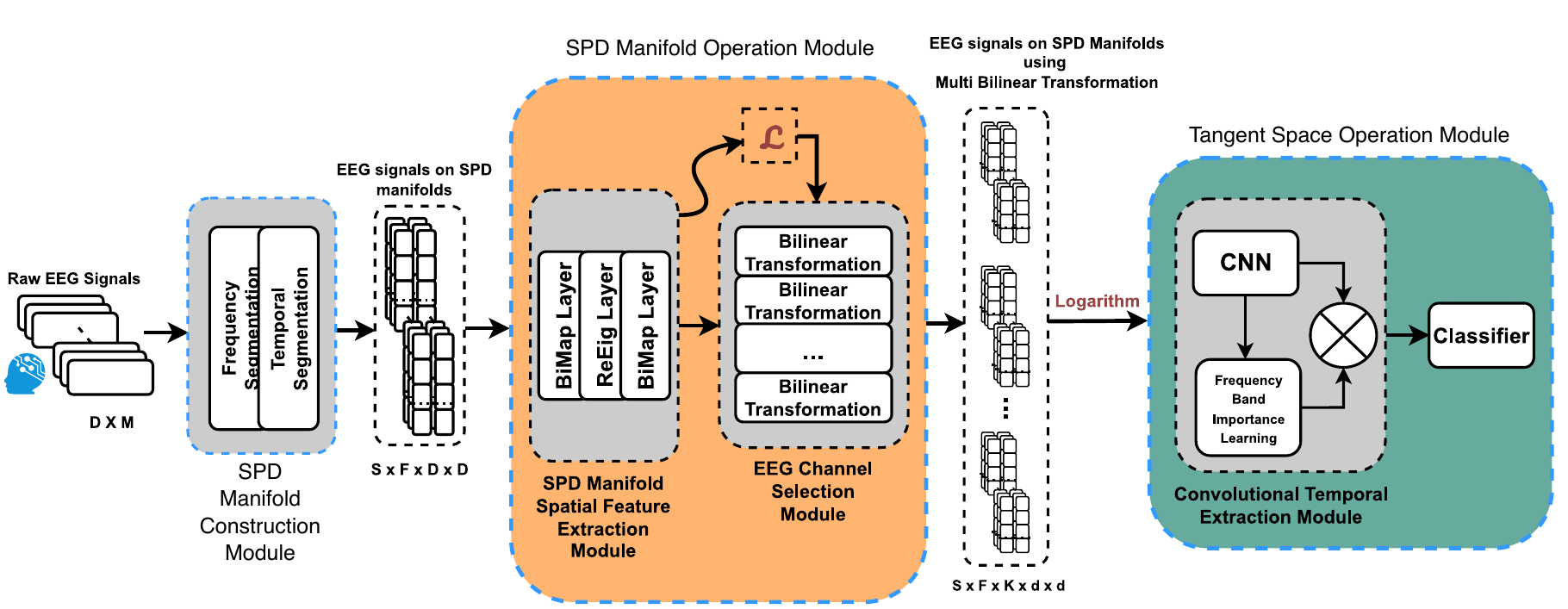}
\caption{Architectural overview of {\tool}}
\label{fig:Architecture}
\end{figure*}

\subsection{Architecture}

As shown in Figure~\ref{fig:Architecture},
the {\tool} prototype is comprises three components, given below. 
As detailed in Section~\ref{sec:hardware}, raw EEG signals gathered by our prototype 
go through these modules sequentially for the final classification output, 
which we elaborate in the following subsections:

\begin{enumerate}

\item SPD manifold construction module:
This module acts as the signal processing
block in the {\tool} prototype. Details are given in Section~\ref{sec:module-1}.

\item SPD manifold operation module:
The function of this module is to
capture inherent spatial patterns in the original Riemannian geometry of EEG signals
and to choose the representative subset of EEG channels.
It is composed of two sub-modules: the SPD manifold spatial feature extraction module and the EEG channel selection module.
We delve into these two sub-modules
in Section~\ref{sec:module-2-1} and Section~\ref{sec:module-2-2}.

\item Tangent space operation module:
In this final module, CNNs are employed to capture the temporal dynamics of EEG signals on the tangent space,
followed by a classifier for the final output. We discuss this module in Section~\ref{sec:module-3}.

\end{enumerate}

\subsection{SPD Manifold Construction Module}
\label{sec:module-1}
The SPD manifold construction module is
responsible for transforming raw EEG data into SPD manifolds,
acting as the signal processing block in the {\tool} prototype.
This module receives multi-channel raw data from our data pre-processing pipeline, where the raw data are segmented both temporally and by frequency.

\begin{itemize}
\item To segment the EEG signals in the frequency domain,
we employ a well-known filter bank technique~\cite{ang2008filter}.
This technique uses a bank of bandpass filters to decompose the raw oscillatory EEG signals
into multiple frequency passbands.
We choose the causal Chebyshev Type II filter to ensure the non-negativity of filter coefficients
and to minimize the filter transition bandwidth~\cite{ju2022tensor}.
\item For temporal segmentation,
we adopt a fixed-interval strategy. Initially,
we divide the EEG signals into short, non-overlapping intervals of equal length.
The length of each time window, denoted as $L$, is determined based on Gabor's uncertainty principle~\cite{gabor1946theory}:
\begin{equation}
   \Delta t \cdot \Delta f \geq \frac{1}{4\pi},
   \label{eq:bilinear transformation}
\end{equation}
where $\Delta t$ is the time width of the signal, and $\Delta f$ is the frequency width of the signal.
Gabor's uncertainty principle states that time and frequency resolutions cannot be simultaneously maximized. This strategy ensures a balance between temporal and frequency resolutions of the EEG signals.
\end{itemize}

After frequency and temporal segmentation,
we create four-dimensional tensors, $T \in \mathbb{R}^{S \times F \times M \times L}$,
by stacking segments. Here, $S$, $F$, $M$, and $L$ denote window slices, filter banks, channels,
and the window length, respectively. These tensors, $T$, help compute spatial covariance matrices,
$X \in \mathbb{R}^{S \times F \times M \times M}$,
from EEG signals on SPD manifolds, serving as inputs for the spatial feature extractor module.

\subsection{SPD Manifold Operation Module: \textit{Feature Extraction}}
\label{sec:module-2-1}

The SPD manifold spatial feature extraction module,
which is the first sub-module of the SPD manifold operation module,
is designed based on BiMap to capture inherent spatial patterns in original Riemannian geometry.
The concept of BiMap was initially introduced in the context of SPDNet~\cite{huang2017riemannian},
primarily for the purpose of transforming the covariance matrix $S$ using the
BiMap operator $W \cdot S \cdot W^T$. This transformation is particularly suited
for SPD matrices because it maintains their SPD properties,
which is crucial for handling biomedical signals such as EEG.

In our work, we extended the traditional application of
the BiMap layer by employing a multi-layer BiMap structure on the SPD manifold. We construct an architecture akin to a multilayer perceptron
utilized in Euclidean spaces for feature extraction.
This approach enables us to effectively extract meaningful features
from the original SPD data while preserving the geometric properties of the data. This is vital for enhancing the performance of BCI systems.
The output is normalized using Riemannian batch normalization.

\subsection{SPD Manifold Operation Module: \textit{Channel Selection}}
\label{sec:module-2-2}

The second sub-module of the SPD manifold operation module is the EEG channel selection module,
which involves identifying and choosing a subset of EEG channels
that effectively capture the spatial patterns and structures
presented in EEG signals that are obtained from the previous module.
The objective is to reduce the dimensionality of the SPD matrix
from $Sym_{++}^{M}$ to $Sym_{++}^{m}$, where $M>>m$, while retaining the most important EEG channels.
As traditional channel selection methods in
Euclidean space~\cite{abdi2022eeg,strypsteen2021end} are not suitable in SPD manifolds,
we developed a geometric-aware bilinear transformation to select critical EEG channels.
The transformation is similar to BiMap from
the RMTSISOM-SPDDR method~\cite{gao2021dimensionality} but adheres to two key properties:

\begin{itemize}
\item Property 1: Preservation of the original SPD data points.
We ensure that data points remain on the SPD manifold after the transformation.
\item Property 2: Maintaining distances.
We ensure that the distances between data points in the tangent space
after the transformation closely approximate their
corresponding geodesic distances on the SPD manifold.
\end{itemize}

We next describe how we construct the objective function
based on these two properties to assist {\tool} in learning this transformation,
thereby obtaining the most crucial EEG channels for specific MI tasks.

\subsubsection{Preservation of Original SPD Data Points}

Ensuring the preservation of the original SPD data points is vital
when transforming tangent vectors on the SPD manifold.
We impose the constraint $\bm {W^T W = I_m}$ to ensure that the transformed tangent vectors remain SPD
and that they can be accurately mapped back to the manifold.

\subsubsection{Maintaining Distances}
\label{sec:airmdistance}

There are two distances:
geodesic distance and Euclidean distance.
The set of SPD matrices $Sym_{++}^{m}$ forms a Riemannian manifold.
Hence, the geodesic distance between two SPD matrices $X_{i}$ and $X_{j}$ under
the affine-invariant Riemannian metric (AIRM)~\cite{pennec2006riemannian},
which is invariant to affine transformations, can be expressed as follows:
\begin{equation}
g_{ij} =  \|\log(X_{i}^{-\frac{1}{2}} X_{j} X_{i}^{-\frac{1}{2}})\|_{F}, \quad 1 \leq i,j \leq N,
\label{eq:geodesic-distance-ij}
\end{equation}
where $\|\cdot\|_F$ denotes the Frobenius norm.
The tangent space is Euclidean space. This allows for the computation of
the Euclidean distance between tangent vectors. Specifically,
for $X_{i}$ and $X_{j}$ from the SPD manifold, they are first mapped to its tangent space
using the LogEig layer, resulting in the tangent vectors $ \log(X_{i})$ and $\log(X_{j})$.
These vectors, once transformed via the BiMap layer
(i.e. $W^{T} \log(X_{i})W$ and $W^{T} \log(X_{j})W$), have their distance computed as:
\begin{equation}
\begin{aligned}
d_{ij} &= \|W^{T}\log(X_{i})W - W^{T}\log(X_{j})W\|_{F} \\
&= \|W^{T}(\log(X{i}) - \log(X_{j}))W\|_{F},
\label{eq:Euclidean Distance}
\end{aligned}
\end{equation}

Let \( G = [g_{ij}] \) and \( D = [d_{ij}] \) represent
the geodesic distance matrix and Euclidean distance matrix, respectively.
To preserve the distances, it is expected that \( G = D \).

\subsubsection{Objective Function for the Channel Selection}

To maintain equality between geodesic and Euclidean distance matrices,
we designed an objective function to learn the optimal
value of $W$ in Eq.~\ref{eq:Euclidean Distance}:
\begin{equation}
\begin{aligned}
\hat{W}
&=  \mathop{\arg\min}\limits_{W} \sum_{i=1}^{N}\sum_{j=1}^{N}(g_{ij} - d_{ij})^{2}\\
&= \mathop{\arg\min}\limits_{W} \|G-D\|^{2}_{F}\ \text{(Property 2)} \\
&s.t.\ W^{T}W=I_{m}\ \text{(Property 1)}
\end{aligned}
\label{Object_1}
\end{equation}

For $\forall{g_{ij}}{\in}G$ and $\forall{d_{ij}}{\in}D$,
$(g_{ij} - d_{ij})^{2} = g_{ij}^{2} + d_{ij}^{2} - 2g_{ij}d_{ij}$.
Given that $g_{ij}$ is expected to have a value very close to $d_{ij}$,
the expression can be approximated
as $g_{ij}^{2} + d_{ij}^{2} - 2g_{ij}d_{ij} \approx g_{ij}^{2} + d_{ij}^{2} - 2d_{ij}^{2} = g_{ij}^{2} - d_{ij}^{2}$.
Then, Equation~(\ref{Object_1}) can be rewritten as
\begin{equation}
\begin{aligned}
\hat{W}
&=  \mathop{\arg\min}\limits_{W} \sum_{i=1}^{N}\sum_{j=1}^{N}(g_{ij}^{2} - d_{ij}^{2})^{2}\\
&= \mathop{\arg\min}\limits_{W} \|G^{2}-D^{2}\|^{2}_{F}\ \\
&s.t.\ W^{T}W=I_{m}\
\end{aligned}
\label{Object_2}
\end{equation}

\begin{theorem}
If we let $A = (-\frac{1}{2}(G^{2}-D^{2}))$,
then $HAH$ results in a positive semi-definite (p.s.d.) matrix, with $H=I_{m}-m^{-1}11^{T}$ representing the centering matrix.
Here, $1=[1, 1, \dots, 1]^{T} \in \mathbb{R}^{m}$, and $I_{m}$ is an $m \times m$ identity matrix.
\label{th:Positive semi-definite property}
\end{theorem}
Theorem~\ref{th:Positive semi-definite property} is derived from the p.s.d. property~\cite{bartlett1947multivariate},
which states the following: let $A = (-\frac{1}{2}z_{ij}^{2})$, where $z_{ij}^{2}$ represents the Euclidean distance between $i$ and $j$,
and then $HAH$ is p.s.d. Here, we let $z_{ij}^{2} {\in}(G^{2}-D^{2})$.
Based on Theorem~\ref{th:Positive semi-definite property}, we can center the objective function:
\begin{equation}
\begin{aligned}
\hat{W} &= \mathop{\arg\min}\limits_{W} \|H(-\frac{1}{2}(G^{2}-D^{2}))H\|^{2}_{F} \\ &=  \mathop{\arg\min}\limits_{W} \|\gamma_{G}-\gamma_{D}\|^{2}_{F}.
\label{eq.T1_W}
\end{aligned}
\end{equation}
Here, $\gamma_{D}=-\frac{1}{2}H D^{2}H$ and $\gamma_{G}=-\frac{1}{2}H G^{2}H$ are both p.s.d.
It can be proven that both $\gamma_{D}$ and $\gamma_{G}$ are centered inner matrices~\cite{bartlett1947multivariate}, and we can obtain
\begin{equation}
d_{ij}^{2} = (\gamma_{D})_{ii} + (\gamma_{D})_{jj} - 2(\gamma_{D})_{ij}.
\label{eq:cosin_law}
\end{equation}
Moreover, based on the definition of the Frobenius inner product, $\hat{W}$ can be further rewritten as
\begin{equation}
\hat{W} = \mathop{\arg\min}\limits_{W}[-2tr({\gamma_{G}\gamma_{D}}^{T})] +  \|\gamma_{G}\|^{2}_{F} + \|\gamma_{D}\|^{2}_{F}.
\end{equation}
Since $d_{ij} = \|W^{T}(\log(X{i}) - \log(X_{j}))W\|_{F}$ (Eq.~\ref{eq:Euclidean Distance}) and we let $k_{ij} = \|X_{i}-X_{j}\|_{F}$,
it follows that $d_{ij} \leq  k_{ij}$.
Also, $\gamma_{D} = -\frac{1}{2}HD^{2}H$ (Eq.~\ref{eq.T1_W}) and $\gamma_{K} = -\frac{1}{2}HK^{2}H$,
where $K=\{k_{ij}\}$. Thus, $ \|\gamma_{D}\|^{2}_{F} \leq  \|\gamma_{K}\|^{2}_{F}$.
Let
\begin{equation}
\hat{W}^{*} = \mathop{\arg\min}\limits_{W}[-2tr({\gamma_{G}\gamma_{D}}^{T})] +  \|\gamma_{G}\|^{2}_{F} + \|\gamma_{K}\|^{2}_{F}.
\end{equation}
Hence, $\hat{W}$ satisfies a specific inequality:
$\hat{W} \leq \hat{W}^{*}$.

This inequality gives us an upper bound for the objective function. Furthermore, since
$\|\gamma_{G}\|^{2}_{F}$ and $\|\gamma_{K}\|^{2}_{F}$
are unrelated to $W$, $\hat{W}$ can be rewritten as
\begin{equation}
\begin{aligned}
\hat{W} &= \mathop{\arg\min}\limits_{W}(-2tr({\gamma_{G}\gamma_{D}}^{T})) \\
& =  \mathop{\arg\min}\limits_{W}(-2\sum_{i=1}^{N}\sum_{i=1}^{N}(\gamma_{D})_{ij}(\gamma_{G})_{ij}.
\end{aligned}
\label{eq:W_1}
\end{equation}
Let $\theta_{i} = Wlog(X{i})W$ and $\theta_{j} = W\log(X_{j}))W$. From Equation~(\ref{eq:cosin_law}), we have
\begin{equation}
2{\gamma_{D_{ij}}} = \|\theta_{i}\|_{F}^{2} + \|\theta_{j}\|_{F}^{2} - \|\theta_{i}-\theta{j}\|_{F}^{2}.
\label{eq:2gamma}
\end{equation}
Substituting Equation~(\ref{eq:2gamma}) into Equation~(\ref{eq:W_1}), we have
\begin{equation}
\begin{aligned}
\hat{W} &= \mathop{\arg\min}\limits_{W}(\|\theta_{i}-\theta{j}\|_{F}^{2}(\gamma_{G})_{ij}). \\
\end{aligned}
\label{eq:W_2}
\end{equation}
Accordingly, the objective function for the EEG channel selection module can be formulated as
\begin{equation}
    \hat{W}_{t+1} = argmax\ tr(W^{T} \mathcal{L} W) \quad s.t. \ W^{T}W=I_{m}.
\end{equation}
Here, $\mathcal{L}$ is defined as follows:
\begin{footnotesize}
\begin{equation}
    \mathcal{L} = -\sum\limits_{i=1}^{N}\sum\limits_{j=1}^{N}\underbrace{(\gamma_{G})_{ij}}_{\mathcal{F}(G)} \underbrace{(log(X_{i}) - log(X_{j}))\hat{W}_{t}\hat{W}^{T}_{t}(log(X_{j}) - log(X_{i}))^{T}}_{\mathcal{F}(D)}.
\label{eq:object function}
\end{equation}
\end{footnotesize}

The expression \(\mathcal{L}\) comprises functions of \(G\) and \(D\),
denoting a unique link between the geometric and Euclidean distances among EEG channels.
By conducting eigenvalue decomposition on \(\mathcal{L}\),
we unveil the importance of the corresponding eigenvectors in the transformation,
employing the indices of the top \(d\) eigenvectors to discern the most significant \(d\) channels.

\subsubsection{Multi-Bilinear Transformation}
The EEG channel selection module picks the top $d$ features from data using eigenvalues,
which may miss crucial information in unselected channels. To address this, we propose the multi-bilinear transformation (MBT),
taking inspiration from Vaswani et al.'s multi-head attention mechanism \cite{vaswani2017attention}.
MBT employs $K$ bilinear transformations to explore distinct spatial correlations among the top $d$ channels,
thus uncovering $K$ diverse spatial correlations. Unlike standard multi-head attention
that concatenates single heads from varying representation subspaces, our mechanism stacks individual signal heads,
forming a multi-channel feature map with each channel depicting a distinct signal head:
\begin{equation}
MBT = Stack(head_1, head_2, ..., head_K).
\label{eq:our mh}
\end{equation}

\subsection{Tangent Space Operation Module}
\label{sec:module-3}
In {\tool}, CNNs are employed to capture the temporal dynamics of EEG signals on the tangent space
by reshaping the output of the EEG channel selection module.
The reshaping is from $\mathbb{R}^{(S \times F \times K) \times m \times m}$
to $\mathbb{R}^{F \times 1 \times S \times (K \times m \times m)}$, excluding the SPD structure.
Here, $F$, $S$, $K$, and $m$ denote
the number of frequency bands,
the number of window slices,
the number of multi-bilinear transformation,
and the dimension of outputs, respectively.
The temporal dynamics are captured using a 2D CNN,
and through Riemannian geometry,
the classification problem transfers from the manifold domain to the Euclidean domain using the logarithm map.
The 2D CNN, with a kernel width of $S \times (K \times m \times m)$,
encompasses all chosen channels in one kernel, minimizing the impact of the spatial position of EEG electrodes.
The CNN is not applied to the frequency band dimension,
acknowledging varying contributions to MI tasks.
A frequency band importance learning block is defined as follows to weigh significant bands:

\begin{equation}
\mathcal{E} = \sigma_{2}(\bm{\omega}{2}(\sigma{1}(\bm{\omega}{1}(\mathcal{F}{sq}(\mathcal{O})))).
\label{eq:Global Node Attention}
\end{equation}
Here, $\mathcal{O}$ is the output of the 2D CNN,
$\omega_{1} \in \mathbb{R}^{m \times \frac{m}{2}}$ and $\omega_{2} \in \mathbb{R}^{\frac{m}{2} \times m}$
represent a two-layer fully connected neural network,
$\sigma{1}$ is a Relu activation function, and $\mathcal{F}_{sq}(\mathcal{O})$ is a squeeze operation~\cite{hu2018squeeze}:

\begin{equation}
\mathcal{F}{sq}(\mathcal{O}) = \frac{1}{m \times m}\sum^{m}_{i=1}\sum^{m}_{j=1}(\mathcal{O}{ij}).
\label{eq:squeeze operation}
\end{equation}

The sigmoid function, denoted $\sigma_{2}$, is used to evaluate the channel importance.
This choice is motivated by the fact that multiple channels can contribute simultaneously and independently to the overall MI classification task,
allowing for a flexible and distributed representation of frequency band importance.
Finally, $\mathcal{O}$ is updated accordingly:
\begin{equation}
\mathcal{O} = \mathcal{E} \times \mathcal{O}.
\end{equation}

\subsection{Discussion of the {\tool} Approach}

The core of our model's innovation lies in its strategic channel selection mechanism,
which diverges from the traditional Tensor-CSPNet approach~\cite{ju2022tensor}
that processes EEG signals across nine disjoint frequency bands.

Tensor-CSPNet primarily focuses on reducing the dimensionality within the SPD manifold. It
typically simplifies the data from high dimensionality $D$ to a lower one $d$ (where $D \gg d$).
This conventional focus often overlooks
a crucial aspect of EEG data: its inherent multi-channel structure.
Our methodology redefines EEG data analysis by not merely simplifying data within channels,
but by strategically selecting the most informative channels from a larger set $M$ down to a more manageable subset $m$ (where $M \gg m$).
This shift from mere dimensionality reduction to a nuanced channel selection paradigm marks a significant advancement in EEG data processing,
enhancing efficiency and reducing complexity without losing essential information.
The contrast with the methodology described by \cite{gao2021dimensionality} is profound.
In that work, the transformation matrix $W$ is two-dimensional, tailored for scaling down data points from $D$ to $d$ within the manifold space.
By contrast, our transformation matrix $W$ is three-dimensional, integrating the channel dimension $M$,
the data length per channel $L$, and the target number of channels $m$.
This approach is not just an expansion in terms of dimensionality;
it represents a foundational innovation in EEG data processing.

Further distinguishing our model is the introduction of a multi-head mechanism,
inspired by transformer models designed to compensate for potential information loss due to the non-selection of certain channels.
This feature enables our model to explore diverse spatial correlations among the selected channels.
By conducting multiple bilinear transformations, the model captures a broader range of signal characteristics,
enhancing its interpretative power and robustness. This aspect represents a significant departure from the architecture of Tensor-CSPNet,
offering a novel way to maintain signal fidelity even with a reduced channel set.

Additionally, our work advances the SPD manifold learning framework for
EEG signal processing by leveraging geometric deep learning techniques tailored to the unique characteristics of EEG data.
While Tensor-CSPNet also operates within a geometric learning context,
our model introduces specific modifications to the SPD manifold construction and operation modules.
These modifications are meticulously designed to accommodate the model's strategic channel selection and multi-head mechanism,
ensuring seamless integration and optimal performance.

\subsection{Prototype Implementation}
\label{sec:hardware}

\begin{figure}[!t]
    \centering
    \begin{subfigure}[b]{0.4\textwidth}
        \centering
        \includegraphics[width=0.42\textwidth]{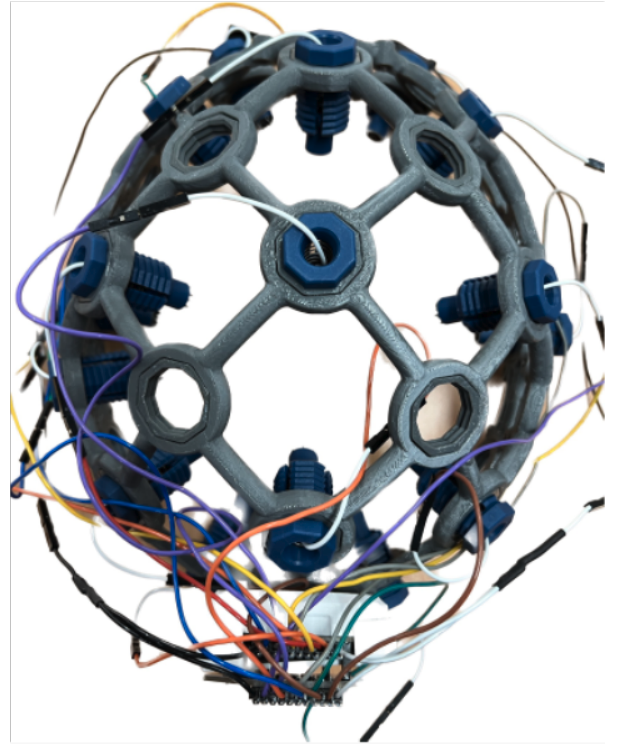}
        \caption{OpenBCI Ultracortex~(Headset-1)}
        \label{fig:openbci}
    \end{subfigure}
    \hspace{20pt}
    \begin{subfigure}[b]{0.4\textwidth}
        \centering
        \includegraphics[width=0.5\textwidth]{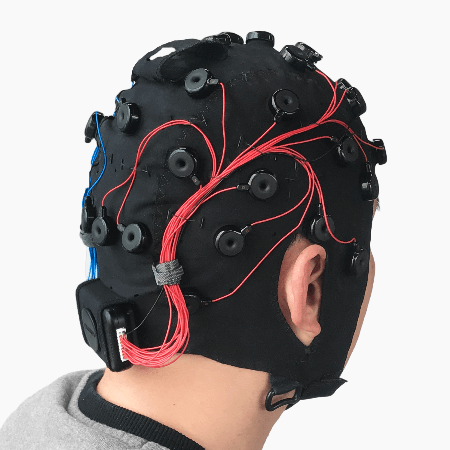}
        \caption{Emotiv EPOC Flex~(Headset-2)}
        \label{fig:flex}
    \end{subfigure}
    
   \caption{Photos of different hardware sets}
   \label{fig:hardware}
\end{figure}

To present the generalization ability of our {\tool} approach,
we implemented two sets of BCI prototypes from different EEG suppliers,
one from OpenBCI (Headset-1),\footnote{https://shop.openbci.com/products/ultracortex-mark-iv}
and
the other
from Emotiv (Headset-2).\footnote{https://www.emotiv.com/product/epoc-flex-saline-sensor-kit/}
The first EEG data collector is the Ultracortex Mark IV 16-channel EEG Cap~\cite{aldridge2019accessible} supplied by OpenBCI (as shown in Figure~\ref{fig:openbci}),
with real-time data collection via the OpenBCI API and a sample rate of 128 Hz.
This headset uses dry, spiky electrodes to penetrate through the hair and make contact with the scalp for EEG signal gathering.
The second EEG data collector is the EPOC Flex supplied by Emotiv (as shown in Figure~\ref{fig:flex}).
Emotiv also provides real-time EEG signal data output through Emotiv Cortex API.
This headset is based on wet-electrode conduction, where the electrodes make contact with the scalp through a soft felt soaked with saline.

As Headset-1 uses dry electrodes,
it facilitates a quicker preparation for experiments.
However, the spiky head of the electrodes induces a tingling sensation and the user is less comfortable.
Meanwhile, the spiky electrodes have a reduced contact area that could lead to lower SNR.
By contrast, although Headset-2 utilizes wet electrodes with saline-soaked felt,
which requires an extended preparation time,
it tends to be more comfortable for the user and provides good contact quality.
It is worth noting that there is variation in the raw EEG data values between these two headsets, which could serve as a test for {\tool}'s generalization.

\section{Experimental Design}
\label{sec:exp}

In this section, we first present the two research questions of this study.
Subsequently, we provide detailed introductions to the experimental setups for investigating these research questions.
We formulated two research questions as follows:

\begin{itemize}

 \item{RQ1:} How does {\tool}'s generalization capability perform on EEG-based motor imagery tasks?
 We performed deep analysis through a comparison of several baseline methodologies to present the capacity of {\tool}.

 \item{RQ2:} How does {\tool} work in a real-world motor imagery setting, with different EEG devices?
 We conducted real-world experiments with the two devices mentioned in Section~\ref{sec:hardware} to test the performance of {\tool}.

\end{itemize}

\subsection{Experimental Setup}

\subsubsection{General Setup}
Training occurred over 100 epochs on a desktop computer
equipped with an NVIDIA GeForce RTX 4090 GPU.
During training, cross-entropy was chosen as the loss function for its simplicity, 
with a batch size of 64 and a learning rate of $1e^{-3}$.
For the inference, {\tool} ran on a desktop equipped with an NVIDIA GeForce RTX 2080Ti GPU,
which processed the EEG data for motor-imagery (MI) classification.

\subsubsection{Setup for RQ1}
\label{sec:rq1}
We evaluated {\tool}'s generalization using two public MI datasets: 
MI-KU~\cite{lee2019eeg} and BCIC-IV-2a~\cite{tangermann2012review}.
MI-KU has 54 participants in a binary-class MI task, 
recorded via 62 electrodes at 1,000 Hz, with two sessions of 200 trials each.
BCIC-IV-2a features nine subjects in a four-class MI-EEG task, 
recorded via 22 electrodes and three EOG channels at 250 Hz,
with training and evaluation sessions comprising six runs of 12 trials for each class, 
totaling 288 trials per subject.
{\tool} was benchmarked against several baseline techniques to evaluate its generalization capabilities: a CSP-based approach (FBCSP~\cite{ang2008filter}),
Riemannian-based approaches (MDM~\cite{tevet2022human}; TSM~\cite{lin1811temporal}; SPDNet~\cite{huang2017riemannian}; Tensor-CSPNet~\cite{ju2022tensor}; and Graph-CSPNet~\cite{ju2023graph}),
and traditional deep learning methodologies (EEGNet~\cite{lawhern2018eegnet}; ConvNet~\cite{liu2023novel}; and FBCNet~\cite{mane2021fbcnet}).

\subsubsection{Setup for RQ2}
\label{sec:rq2}
To evaluate the real-world performance of {\tool},
we designed a two-step experimental procedure:
(1) an EEG data collection step, and
(2) the {\tool} inference step.
During the experiment, the participants were required to sit 1.5 meters back from a 27-inch monitor.
They were cued by a visual and auditory signal to start an imagery task,
and then performed the required mental execution of movements continuously until the visual cue disappeared (around 10 seconds).

\begin{figure}[t]
    \centering
    \begin{subfigure}{0.24\textwidth}
        \centering
        \includegraphics[width=\textwidth]{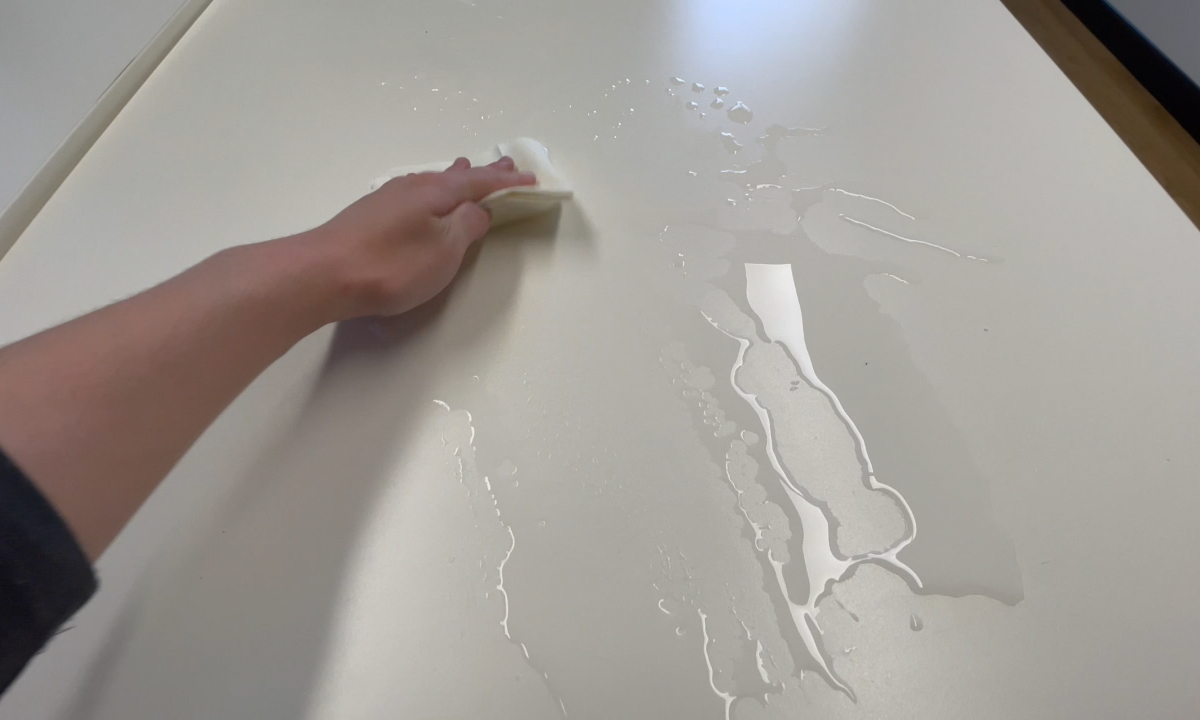}
        \caption{Left Hand}
        \label{fig:left_1}
    \end{subfigure}
    \hfill
    \begin{subfigure}{0.24\textwidth}
        \centering
        \includegraphics[width=\textwidth]{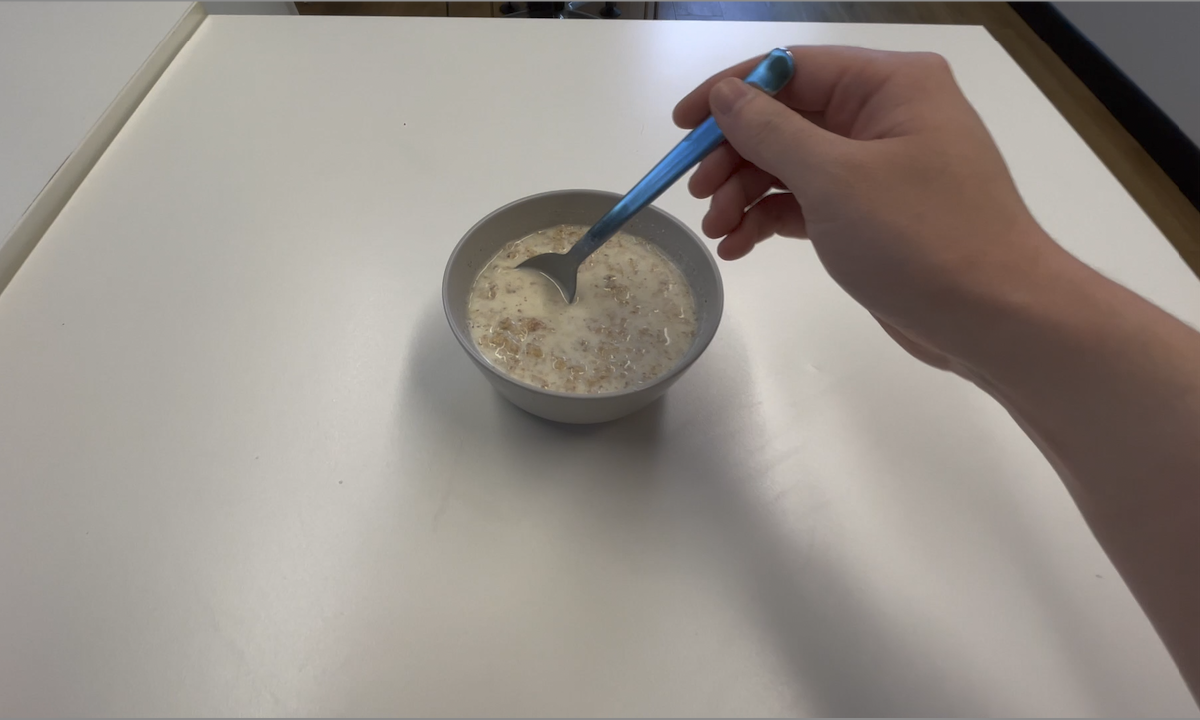}
        \caption{Right Hand}
        \label{fig:right_1}
    \end{subfigure}
    \hfill
    \begin{subfigure}{0.24\textwidth}
        \centering
        \includegraphics[width=\textwidth]{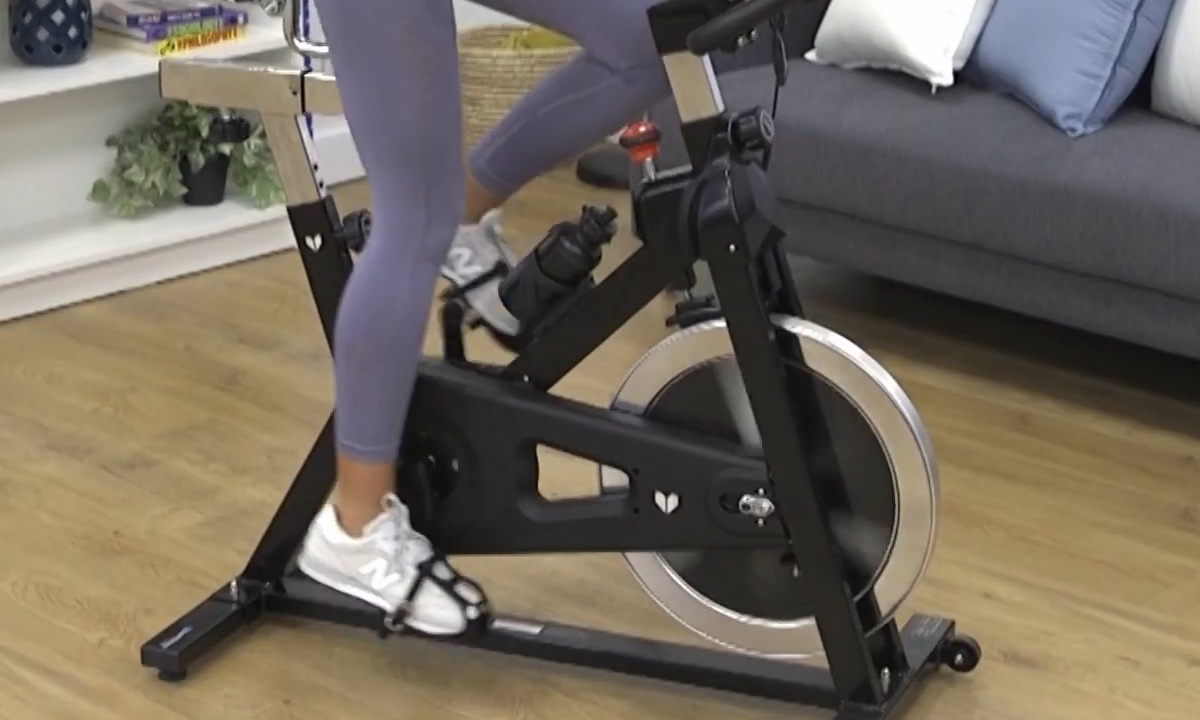}
        \caption{Both Feet}
        \label{fig:leg_1}
    \end{subfigure}
    \hfill
    \begin{subfigure}{0.24\textwidth}
        \centering
        \includegraphics[width=\textwidth]{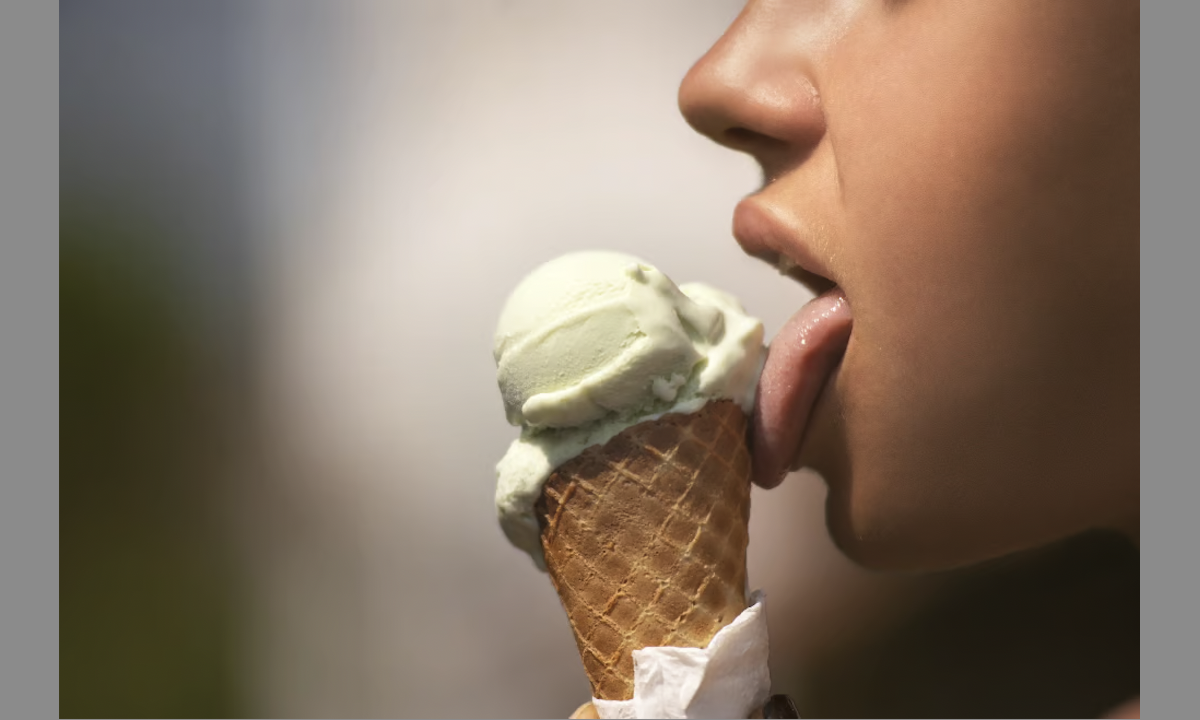}
        \caption{Tongue}
        \label{fig:tongue_1}
    \end{subfigure}

    \caption{Images signifying motor-imagery tasks}
    \label{fig:task-images}
\end{figure}

\textbf{EEG Data Collection Step:}
As we set wheelchairs as the moving target, at least four commands
were required for the movement: Forward, Stop,
Left Rotation, and Right Rotation.
The direct thoughts of these movements are generated in a deep layer of brain
(as mentioned in Section~\ref{sec:eeg-wheelchair}),
making the data difficult to extract.
Our workaround is to encode the
EEG-MI task classes to the controlling command of the wheelchair.
The four classes of MI tasks we selected---imagining the movements of the Left Hand, Right Hand, Both Feet, and Tongue---
have previously proven their efficacy at stimulating different brain regions~\cite{pfurtscheller2006mu}.
These four movements are also easily encoded to wheelchair movements.
However, for these four classes there are thousands of possibilities of specific activities.
Inspired by \citet{rassam2024competing},~\citet{tsuda2014analysis}, and~\citet{wilson2023feasibility},
and to
reduce the variance in our real-world scenario experiment,
we selected one specific activity for each of the motor imagery task (as shown in Figure~\ref{fig:task-images}):

\begin{itemize}
    \item For the Left Hand, we chose the movement ``wiping a desk"
    \item For the Right Hand, we chose the movement ``using a spoon for food"
    \item For Both Feet, we chose the movement ``riding a spin bike"
    \item For the Tongue, we chose the movement ``licking ice cream"
\end{itemize}

We recruited volunteers to help with the EEG data collection.
The participants were required to perform the 4 mentioned motor imagery tasks.
The data collection for each participant included six runs,
each with all four tasks (Figure~\ref{fig:task-images})
where each specific task was repeated 10 times.
Thus, there were 60 repetitions per specific task in total.
The procedure was conducted on prototype Headset-1.
We recruited 20 volunteers equally split
by gender. They were 20--25 years old (mean age: 23 years).
To ensure that the participants were genuinely following cues and performing the imagery tasks,
the order of the four tasks within each run was randomized.
Thus, it was not possible for the participants to find out the presentation pattern of cues.
There was a 10-minute break after each run to
provide enough rest time for the participants for better imagery results and to mitigate fatigue.
For prototype Headset-2, we gathered extra data from two additional participants
(one male and one female) wearing Headset-2,
The data collection step included two runs with 40 trials each,
totaling 20 repetitions per task. The gathered data were used to fine-tune the model trained by the Headset-1 data.

\textbf{{\tool} Inference Step:}
Three independent individuals who did not participate in the data collection process
were assigned to perform the four random motor-imagery tasks in random order.
The inference step included one run 15 times for each specific task per participant,
leading to 180 commands in total.

\section{Experimental Results}
\label{sec:result}

\subsection{RQ1 Result Analysis: {\tool} Generalization Performance vs. State-of-the-art Methodologies}

\begin{table*}[!tb]
	\centering
\caption{Comparative Analysis of Subject-Specific Accuracy and Standard Deviations in MI-KU and BCIC-IV-2a Datasets}
\scalebox{0.8}{ 
    \begin{tabular}{lccc|ccc}

    \toprule
    \toprule
	    & \multicolumn{3}{c}{MI-KU} & \multicolumn{3}{c}{BCIC-IV-2a} \cr
     
    \cmidrule(lr){1-7}
        & CV       (S1) & CV       (S2) & Holdout (S1 $\rightarrow$ S2) & CV        (T) & CV        (E) & Holdout (T $\rightarrow$ E) \cr
        & Acc \% (Std.) & Acc \% (Std.) & Acc \%                 (Std.) & Acc \% (Std.) & Acc \% (Std.) & Acc \%               (Std.) \cr
        
    \cmidrule(lr){1-7}
		FBCSP~\cite{ang2008filter}         & 64.33 (15.43) & 66.20 (16.29) & 59.67 (14.32) & 71.29 (16.20) & 73.39 (15.55) & 66.13 (15.54) \cr
    \cmidrule(lr){1-7}
		EEGNet~\cite{lawhern2018eegnet}    & 63.35 (13.20) & 64.86 (13.05) & 63.28 (11.56) & 69.26 (11.59) & 66.93 (11.31) & 60.31 (10.52) \cr
		ConvNet~\cite{liu2023novel}        & 64.21 (12.61) & 62.84 (11.74) & 61.47 (11.22) & 70.42 (10.43) & 65.89 (12.13) & 57.61 (11.09) \cr
        FBCNet~\cite{mane2021fbcnet}       & 73.36 (13.71) & 73.68 (14.97) & 67.74 (14.52) & 75.48 (14.00) & 77.16 (12.77) & 71.53 (14.86) \cr
    \cmidrule(lr){1-7}
		MDM~\cite{tevet2022human}          & 50.47 (8.63)  & 51.93 (9.79)  & 52.33 (6.74)  & 62.96 (14.01) & 59.49 (16.63) & 50.74 (13.80) \cr
		TSM~\cite{lin1811temporal}         & 54.59 (8.94)  & 54.97 (9.93)  & 51.65 (6.11)  & 68.71 (14.32) & 63.32 (12.68) & 49.72 (12.39) \cr
        SPDNet~\cite{huang2017riemannian}  & 57.88 (8.68)  & 58.88 (8.68)  & 60.41 (12.13) & 65.91 (10.31) & 61.16 (10.50) & 55.67 (9.54)  \cr
		Tensor-CSPNet~\cite{ju2022tensor}  & 73.28 (15.10) & 74.16 (14.50) & 69.50 (15.15) & 75.11 (12.68) & 77.36 (15.27) & 73.61 (13.98) \cr
        Graph-CSPNet~\cite{ju2023graph}    & 72.51 (15.31) & 74.44 (15.52) & 69.69 (14.72) & 77.55 (15.63) & 78.82 (13.40) & 71.95 (13.36) \cr
    \cmidrule(lr){1-7}
        ${\tool}^{(CHL: 20, Head: 4)}$     & 72.56 (14.33) & 73.90 (14.7)  & 69.10 (14.52) & 76.95 (12.41) & 77.13 (12.17) & 73.69 (12.14) \cr
    \bottomrule
    \bottomrule
    
    \end{tabular}
}
\label{tab:SOTA Comparison}
\vspace{-5pt}
\end{table*}

\textbf{Comparison with Other State-of-the-art Methods:}
To assess the generalization capabilities of {\tool},
we compared it with popular methods of handling MI tasks on
the MI-KU~\cite{lee2019eeg} and BCIC-IV-2a~\cite{tangermann2012review} datasets.
MI-KU contains two sessions:
Session-1~(S1) and Session-2~(S2).
Similarly,
the BCIC-IV-2a dataset has a training session~(T) and an evaluation session~(E).
Thus, we adopted the same evaluation methods as in~\cite{ju2022tensor}.
Specifically,
we conducted a comparison of two scenarios:
(1) a cross-validation scenario, which used a 10-fold-cross-validation (CV) technique on each session (S1, S2, T, and E);
and (2) a holdout scenario, where the models were trained in one session (S1 or T) and evaluated in another session (S2 or E).
The performance metrics of the various methodologies are detailed in Table~\ref{tab:SOTA Comparison}.

Conventional deep learning (DL) methods for MI-EEG classification
rely on leveraging the spatiotemporal frequency characteristics of EEG signals.
Both EEGNet and ConvNet utilize these patterns,
achieving performance metrics comparable to FBCSP~\cite{ang2008filter},
a known strategy for extracting spatiotemporal frequency patterns.
This performance equivalence suggests that integrating
any two components can markedly enhance the classification process.
FBCNet, drawing on spatiotemporal frequency patterns,
surpasses EEGNet and ConvNet across all scenarios,
primarily attributed to the employment of bandpass filters adept at capturing frequency information.
For MI-EEG classification using Riemannian-based models, 
AIRM~\cite{pennec2006riemannian} is a distance measure on the Riemannian manifold of $S^n_{++}$,
the space of $n \times n$ SPD matrices (see Section~\ref{sec:airmdistance}). The MDM approach calculates this AIRM distance using geodesic distance, while TSM calculates it using projected SPD matrices.

As Table~\ref{tab:SOTA Comparison} displays,
their performance on the MI-KU dataset is almost the same as random guessing,
though a slight enhancement is observed on the BCIC-IV-2a dataset.
Despite SPDNet introducing a fresh approach for deep non-linear learning on SPD matrix Riemannian manifolds,
it trails all other Riemannian-based models and DL methodologies in Table~\ref{tab:SOTA Comparison}.
The outcomes of MDM, TSM, and SPDNet underline the limitations of
employing geometric quantities of SPD manifolds for high-level classification features.
A notable performance improvement is attained with Riemannian-based models in Tensor-CSPNet and Graph-CSPNet,
especially Graph-CSPNet, which marginally excels in nine out of 11 scenarios.
This superior performance is largely credited to its refined time-frequency segmentation technique,
offering more accurate EEG signal characterization.

\begin{table}[!t]
\begin{minipage}[t]{.49\textwidth}
\centering
\captionsetup{justification=centering}
\caption{Performance Comparison and\\ Parameter Efficiency in BCIC-IV-2a of\\ Tensor-CSPNet, Graph-CSPNet, and {\tool}}
\label{tab:rq1 performance comparison}
\scalebox{0.91}{
    \begin{tabular}{lcc}
        \toprule
        \toprule
        Method & Parameters ($1K$) & Acc \% (Std.) \cr
        \cmidrule(lr){1-3}
        Graph-CSPNet & 169.4 & \textbf{77.55 (15.63)}   \cr
        {\tool}      & \textbf{100.5} & 76.95 (12.41)   \cr
        Tensor-CSPNet & 232.3  & 75.11 (12.68)        \cr
        \bottomrule
        \bottomrule
    \end{tabular}
}
\end{minipage}
\hfill
\begin{minipage}[t]{.49\textwidth}
\centering
\captionsetup{justification=centering}
\caption{Impact of Channel (CHL) Selection on\\ Model Performance in One and Four-Head Settings\\~}
\label{tab:channel impact}
\scalebox{0.8}{
                \begin{tabular}{ccc}
                \toprule
                \toprule
                \# of CHLs & Acc \% (Std.) - 1 Head & Acc \% (Std.) - 4 Heads \cr
                \cmidrule(lr){1-3}
                20 & \textbf{73.93 (15.67)} & \textbf{76.95 (13.17)} \cr
                15 & 73.37 (11.62) & 75.74 (13.87) \cr
                10 & 69.83 (14.86) & 73.74 (11.18) \cr
                5  & 65.02 (9.35) & 72.18 (9.29) \cr
                \bottomrule
                \bottomrule
            \end{tabular}
}
\end{minipage}
\vspace{-5pt}
\end{table}

In Table~\ref{tab:SOTA Comparison},
although the accuracy of {\tool} is not the highest,
it is still competitive with no noticeable gap.
On the other hand, {\tool} requires
many fewer parameters,
a milestone reached by its capacity to discern and select the most pivotal channels.
This not only minimizes the learnable parameter size but also enhances its efficacy at pinpointing the most crucial frequency bands.
Table~\ref{tab:rq1 performance comparison} presents a performance comparison on the BCIC-IV-2a dataset of Tensor-CSPNet, Graph-CSPNet, and {\tool}.
{\tool} displays an accuracy increment of around $2\%$ over Tensor-CSPNet ($76.95\%$ vs. $75.11\%$),
while cutting down the parameter count by more than $130K$ ($100.5K$ vs. $232.3K$).
Although {\tool}'s accuracy is slightly below Graph-CSPNet's by about $1\%$ ($76.95\%$ vs. $77.55\%$),
it achieves this outcome with a significant parameter reduction of about $70K$ ($100.5K$ vs. $169K$).
Thus, with a streamlined parameter count, {\tool} approaches state-of-the-art results.

In addition to accuracy, the standard deviation analysis highlights the robustness of {\tool}.
Evaluating the standard deviations provides insights
into the consistency of a model's performance across different subjects and conditions.
On the BCIC-IV-2a dataset, {\tool} achieves a standard deviation of 12.41,
which is closely aligned with the 12.68 standard deviation of Tensor-CSPNet,
demonstrating similar levels of consistency.
Meanwhile, Graph-CSPNet shows a higher standard deviation of 15.63.
This comparison illustrates that {\tool} not only achieves competitive accuracy
but also exhibits less variance in its performance,
suggesting that it is less sensitive to fluctuations in the data.
This stability is crucial for practical applications where consistency and reliability are key,
as it ensures that {\tool} can deliver reliable results across diverse conditions and subjects.

\textbf{Influence of Varied Channel and Head Selection on the Efficacy of {\tool}:}

To further evaluate the model's performance,
we conducted additional ablation studies and visualizations based on the BCIC-IV-2a dataset.
According to Table~\ref{tab:channel impact},
it aligns well with the common sense that the performance of the model improves as the number of EEG channels used increases.
Interestingly, we also observed that as the number of selected channels decreases,
the model's accuracy did not show a significant decline.
Notably, even with a decrease in the number of channels from $20$ to $5$, the model still achieved comparatively high accuracy.
This finding highlights the effectiveness of our channel selection module at accurately identifying the important channel contributions to the MI task.

Our proposed channel selection mechanism adeptly mitigates computational complexity by strategically omitting less significant channel information. 
This in turn reduces the dimensionality of the SPD matrix, while maintaining the model's precision. 
Despite the loss of certain non-essential channel information, 
we successfully optimized the model’s performance.
Drawing inspiration from the ingenious design of multi-head attention, 
we crafted a multi-head-based subspace mapping method. 
While this approach indeed increases the number of heads, 
it allows for a proportional reduction in the size of each head---that is, 
in the dimension of the respective subspaces.

\begin{figure}[!t]
\centering
\includegraphics[width=0.55\textwidth]{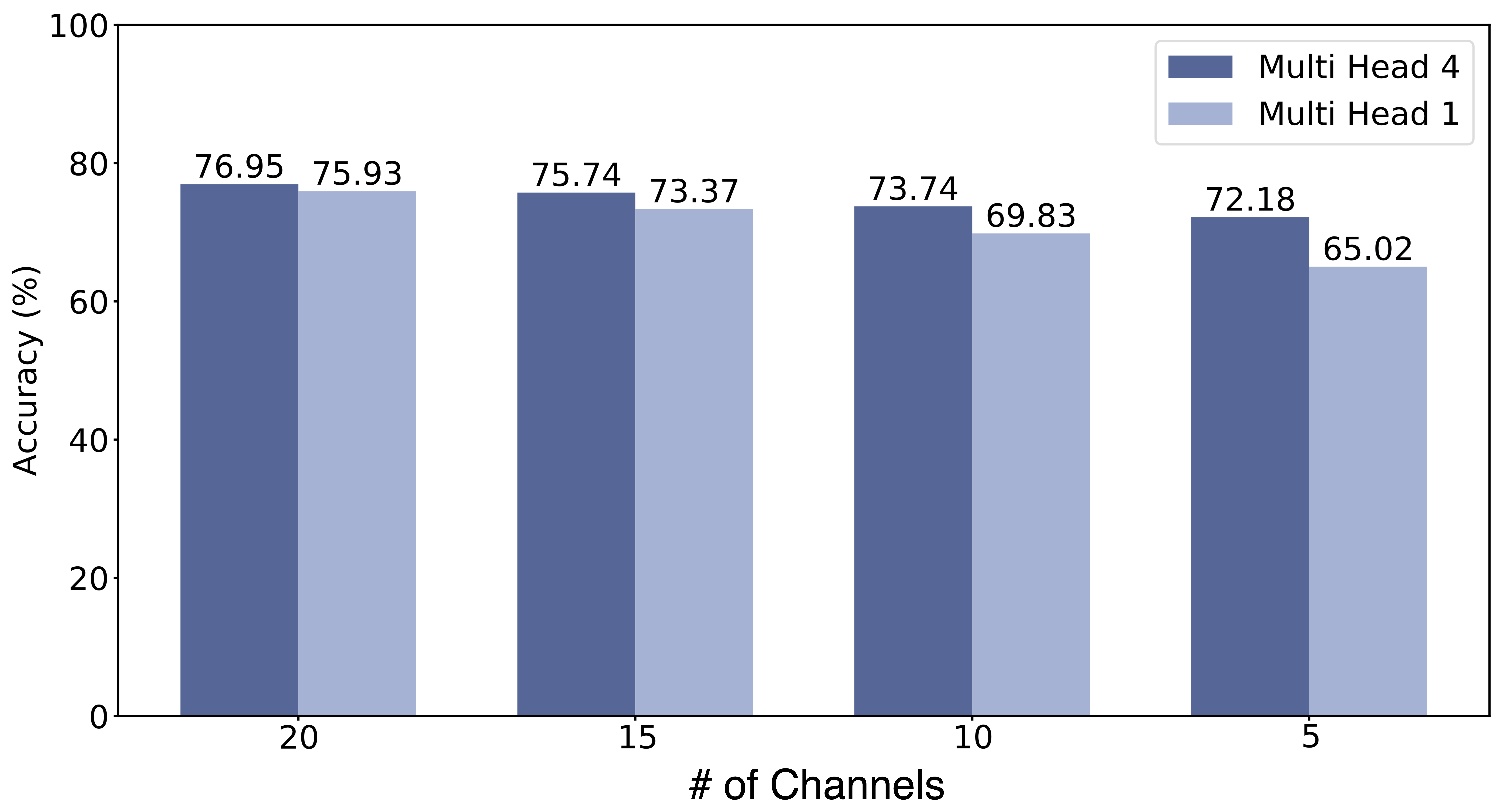}
\caption{Evaluating the performance differential across various channels selected in the one- and four-head settings}
\label{fig: pre and post the application of the multi-head solution}
\end{figure}

Figure~\ref{fig: pre and post the application of the multi-head solution}
provides a comparative analysis of the performance of various models
with different number of channels and with two head selections (one vs. four).
The analysis reveals a consistent improvement in the performance of models that employ multiple heads (namely, four heads).
Interestingly, when all $20$ channels are employed, the performance boost observed post-application of the multi-head solution is less noticeable.
This implies that the use of the full channel spectrum can encapsulate most of the pertinent information necessary for the classification of MI tasks,
whether the multi-head is utilized or not.
However, when only the most critical $K$ channels are used for MI tasks (such as when 15, 10, and 5 channels are selected),
the multi-head solution serves as a compensatory mechanism to account for the loss of less essential channel information.
The findings indicate that with fewer channels selected, there is greater potential to improve the accuracy.
For instance, when $15$ channels are selected, the multi-head solution boosts accuracy by $2\%$,
while a significant improvement of over $7\%$ occurs when only $5$ channels are selected.
These results demonstrate the multi-head solution's ability to enhance the model's learning efficiency
and representation of the complex interplay between the manifold and its tangent space mapping,
thus improving the overall performance.

Table~\ref{tab:comparison across different numbers of channels}
and
Table~\ref{tab:comparison across different numbers of heads}
provide comparisons of
the cost and accuracy achieved across different numbers of channels and heads. 
With a constant number of heads (four),
as shown in Table~\ref{tab:comparison across different numbers of channels},
reducing the channel number (from $20$ to $5$ channels) significantly decreases the parameters (from $100.5K$ to $13.4K$),
yet the accuracy decreases only slightly (from $76.95\%$ to $72.18\%$).
This indicates that our proposed model can effectively select the most critical channels representing useful EEG signals.
Conversely, with a constant number of channels, as the number of heads increases (from $1$ to $12$),
accuracy continues to increase, albeit at a decelerating rate (as shown in Table~\ref{tab:comparison across different numbers of heads}).
A balancing point can be found at $(15, 4)$, where the accuracy increase begins to slow, and the parameter size is relatively low.
This analysis provides insights on how to balance model performance between cost and accuracy.

\begin{table}[ht]
\begin{minipage}[t]{.49\textwidth}
\centering
\captionsetup{justification=centering}
\caption{Comparison Across Different Numbers of Channels}
\scalebox{0.9}{
	\begin{tabular}{lcc}
		\toprule
  \toprule
		\# of Channels & Parameters ($1K$)  & Accuracy \% (Std.)      \cr
		\cmidrule(lr){1-3}
		20     & 100.5     & 76.95 (13.17)       \cr
		15     & 33.0      & 75.74 (13.87)       \cr
		5      & 13.4      & 72.18 (9.29)        \cr
		\bottomrule
  \bottomrule
	\end{tabular}
	}
\label{tab:comparison across different numbers of channels}
\end{minipage}
\hfill
\begin{minipage}[t]{.49\textwidth}
\centering
\caption{Comparison Across Different Numbers of Heads}
\scalebox{0.9}{
	\begin{tabular}{lcc}
		\toprule
  \toprule
		\# of Heads   & Parameters ($1K$)   & Accuracy \% (Std.)   \cr
		\cmidrule(lr){1-3}
		12      & 138.0      & 75.91 (14.53)     \cr
		8       & 124.2      & 75.85 (15.21)     \cr
		4       & 65.4       & 75.74 (12.39)     \cr
        2       & 36.0       & 73.74 (11.91)     \cr
        1       & 21.3       & 72.18 (13.64)     \cr
		\bottomrule
  \bottomrule
	\end{tabular}
	}
\label{tab:comparison across different numbers of heads}
\end{minipage}
\vspace{-5pt}
\end{table}

\begin{figure}[!tb]
\centering
\captionsetup{justification=centering}
\captionsetup[figure]{skip=2pt}
\begin{minipage}[t]{.465\textwidth}
\centering
\includegraphics[width=\textwidth]{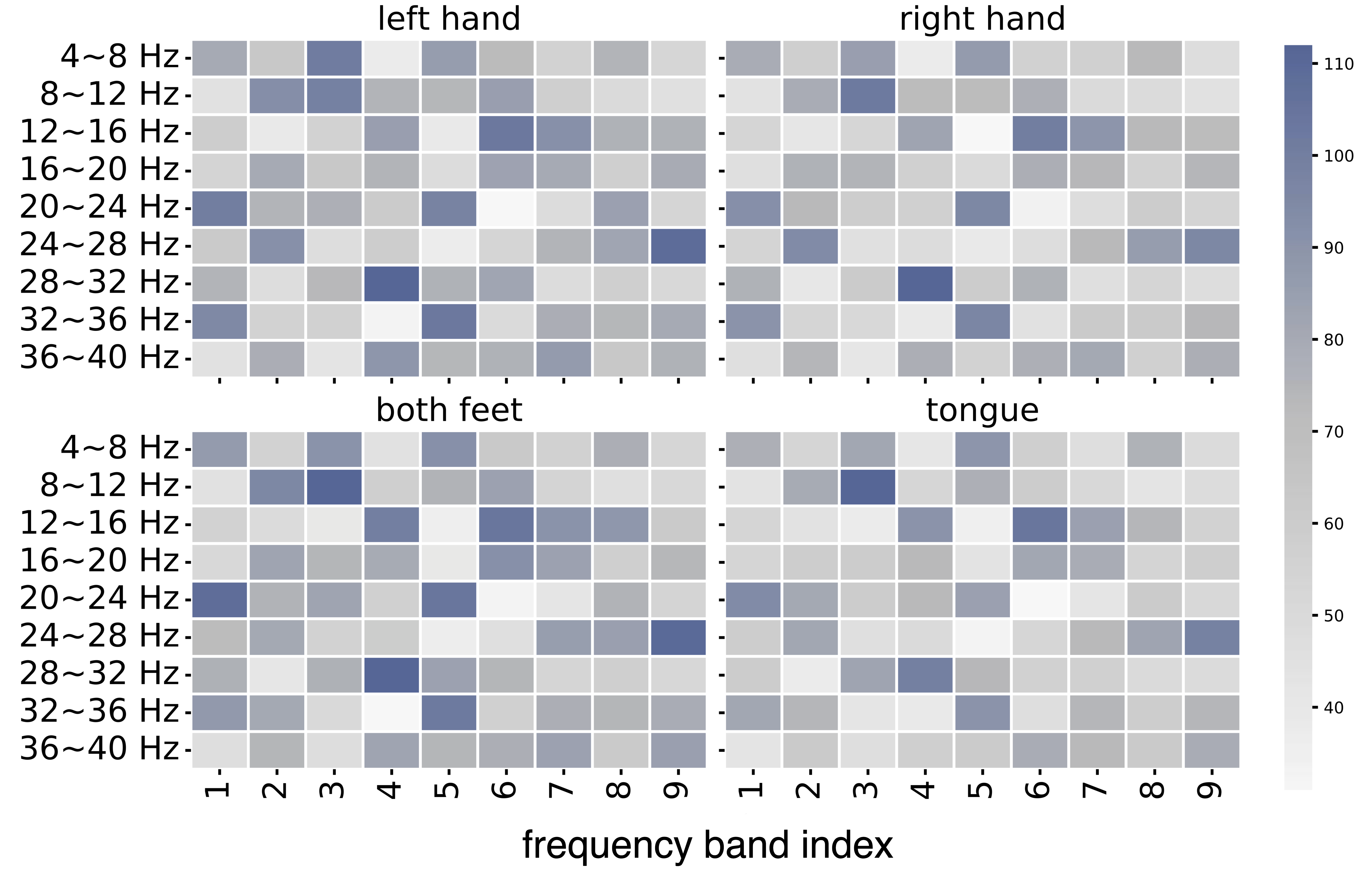}
\caption{Significance heatmap of frequency bands across four MI tasks}
\label{fig:frequence band importance}
\end{minipage}
\hfill
\begin{minipage}[t]{.515\textwidth}
\centering
\includegraphics[width=\textwidth]{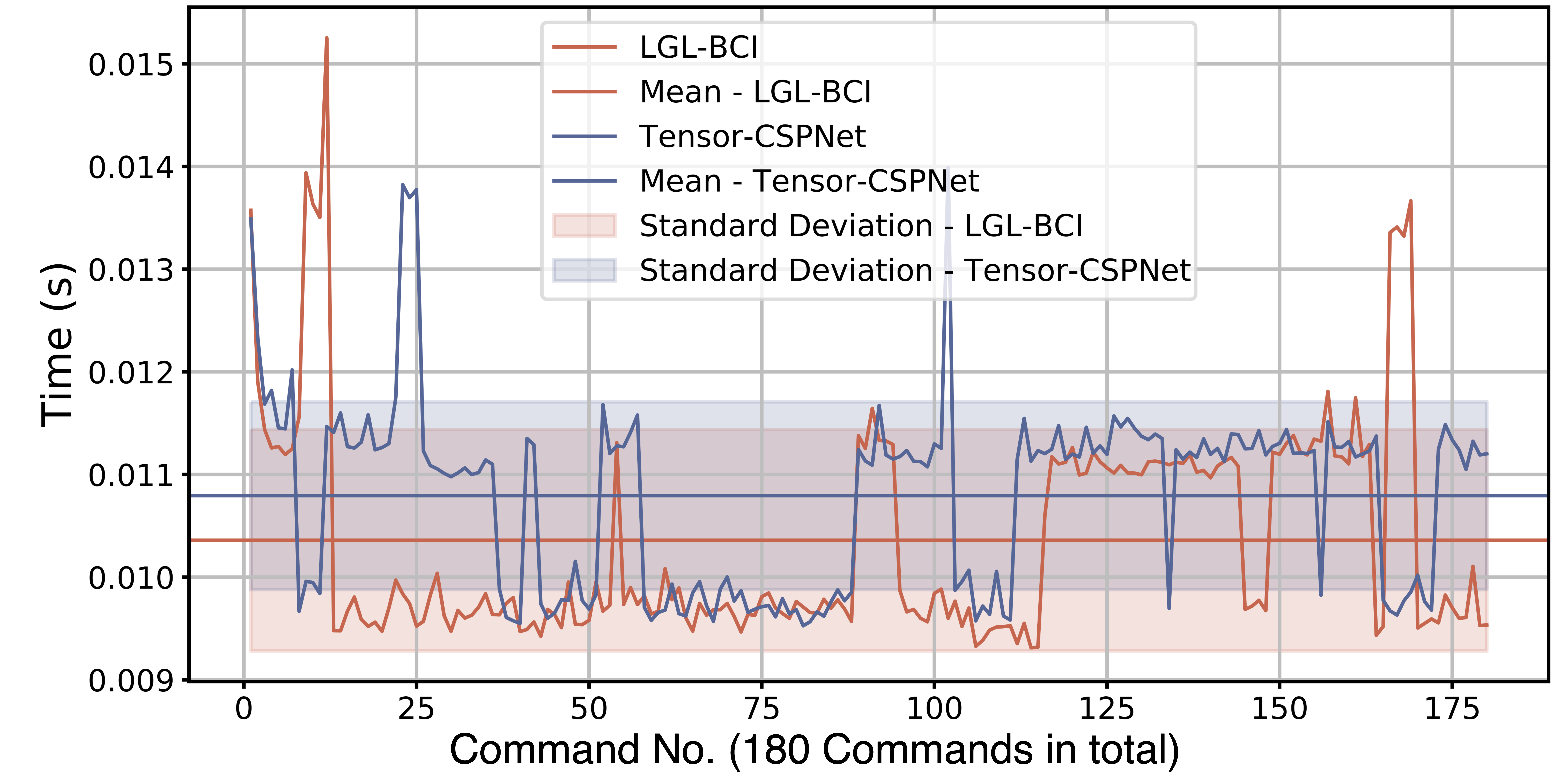}
\caption{Comparative analysis of inference speed: {\tool} vs. Tensor-CSPNet}
\label{fig:Inference Speed}
\end{minipage}
\vspace{-5pt}
\end{figure}

\textbf{Frequency Band Importance Learning:}
Through the study of nine frequency bands,
we discern the significance of these bands across the four MI tasks,
as illustrated in Figure~\ref{fig:frequence band importance}.
In general terms, the importance distribution of these frequency bands shows a remarkable similarity across the four MI tasks.
In this context, Index 1 signifies the frequency band(s) that the model customarily identifies as most vital.
Conversely, Index 9 denotes the frequency band(s) the model typically views as
the
least significant.
A deeper analysis of the chart content leads to the following findings:

\begin{itemize}
\item[$(1)$]
In the Left Hand task, the model generally focuses on the 20--24 Hz and 32--36 Hz bands at Index 1, along with the 24--28 Hz band at Index 2.
A similar pattern is observable in the Right Hand task, which makes sense considering that both tasks involve the imagined motion of a hand.

\item[$(2)$]
For the Both Feet and Tongue tasks,
the model typically assigns more focus to the 20--24 Hz band.
At Index 2, the model dedicated to the Both Feet task usually gives more attention to the 8--12 Hz band,
while the model for the Tongue task does not typically show notable attention to any specific frequency band.

\item[$(3)$]
At Index 3, the model for the Tongue task often gives the most attention to the 8--12 Hz band out of all indices from 1 to 20.
This suggests that for most samples, the 8--12 Hz band is usually seen as particularly informative, even if it is not the most important band (Index 1).
Indeed, this observation at Index 3 is not limited to the Tongue task.
Across all four MI tasks, the 8--12 Hz band is often considered important at Index 3.

\end{itemize}

By studying the significance of frequency bands,
the performance of our model typically improves by approximately $2\%$,
while also generally enhancing the model's interpretability.

\subsection{RQ2 Result Analysis: Performance of {\tool} Under a Real-World Scenario}

Currently, there are no international standards, nor even de facto standards, on the market.
This situation inevitably leads to differences in the output data between products from different suppliers,
due to different designing, manufacturing, and software support.
To explore the performance of {\tool} under this real-world circumstance, we
conducted
an experiment on
the two headsets we built in Section~\ref{sec:hardware},
as the values of raw EEG data provided by Emotiv EPOC Flex differ drastically from those provided by OpenBCI Ultracortex.

To gain comparative insights, we evaluated the performance of {\tool} against the results obtained from Tensor-CSPNet.
Tensor-CSPNet, credited as the trailblazer of the geometric deep learning paradigm in the field of MI-EEG classification,
skillfully utilizes deep neural networks in tandem with the SPD manifolds
to sequentially unravel EEG patterns across the spectrum of frequency, space, and time domains.
Given that {\tool} aligns with a similar signal preprocessing approach and is also architectured based on the SPD manifold,
Tensor-CSPNet is a fitting benchmark to evaluate the prowess of {\tool} in the same scenarios.
As seen in Table~\ref{tab:rq2 performance comparison},
{\tool} surpasses Tensor-CSPNet in terms of accuracy, achieving $82.54\%$ compared to Tensor-CSPNet's $62.22\%$.
Remarkably, it accomplishes this with fewer parameters ($64.9K$ versus Tensor-CSPNet's $183.7K$).

We also conducted comparisons of {\tool} and scaled-down versions of the baseline models with similar parameter counts.
We used L1 unstructured pruning as the scaling-down method following \cite{frankle2018lottery},
with the remaining parameters close to the amount in {\tool}.
The results are reported in Table~\ref{tab:Performance Comparison similar params}.
From these results, we can see that {\tool} surpasses the widely applied scaling-down methodology.

Furthermore, we examined the inference speed of both {\tool} and Tensor-CSPNet.
As illustrated in Figure~\ref{fig:Inference Speed},
compared with Tensor-CSPNet,
{\tool} reduces the time cost by up to 4 ms and about 0.5 ms on average.
Thus, {\tool} infers more quickly,
considering
that the average inference time is within the range of 11 ms.
It is worth noting that the speed difference here is not noticeable, due to strong power of desktop-level GPUs. The difference using embedded devices would be more noticeable.
Our findings reveal that
with about $65\%$ fewer parameters, about $20\%$ more accuracy, and faster inference speed,
{\tool} provides a better experience than Tensor-CSPNet
and is more suitable for real-world applications.

\begin{table}[!ht]
\begin{minipage}[t]{.49\textwidth}
\centering
\captionsetup{justification=centering}
\caption{Comparison of Performance\\ and Parameter Efficiency}
\label{tab:rq2 performance comparison}
\scalebox{0.9}{
    \begin{tabular}{lcc}
        \toprule
        \toprule
        Method & Parameters ($1K$) & Accuracy \% (Std.) \cr
        \cmidrule(lr){1-3}
        {\tool}&    \textbf{64.9}       & \textbf{82.54 (4.36)}        \cr
        Tensor-CSPNet&183.7  &62.22 (7.69)         \cr
        Graph-CSPNet&151.2   &53.10 (5.01)         \cr
        \bottomrule
        \bottomrule
    \end{tabular}
}
\end{minipage}
\hfill
\begin{minipage}[t]{.49\textwidth}
\centering
\captionsetup{justification=centering}
\caption{Comparison of Performance\\ Under Similar Parameter Settings}
\label{tab:Performance Comparison similar params}
\scalebox{0.9}{
    \begin{tabular}{lcc}
        \toprule
        \toprule
        Method & Parameters ($1K$) & Accuracy \% (Std.) \cr
        \cmidrule(lr){1-3}
        {\tool}       & \textbf{64.9} & \textbf{82.54 (4.36}) \cr
        Tensor-CSPNet & 67.8 & 49.05 (6.34) \cr
        Graph-CSPNet  & 66.5 & 45.23 (3.76) \cr
        \bottomrule
        \bottomrule
    \end{tabular}
}
\end{minipage}
\vspace{0pt}
\end{table}

\begin{figure}[t]

    \centering
    \begin{subfigure}{0.325\textwidth}
        \centering
        \includegraphics[width=\textwidth]{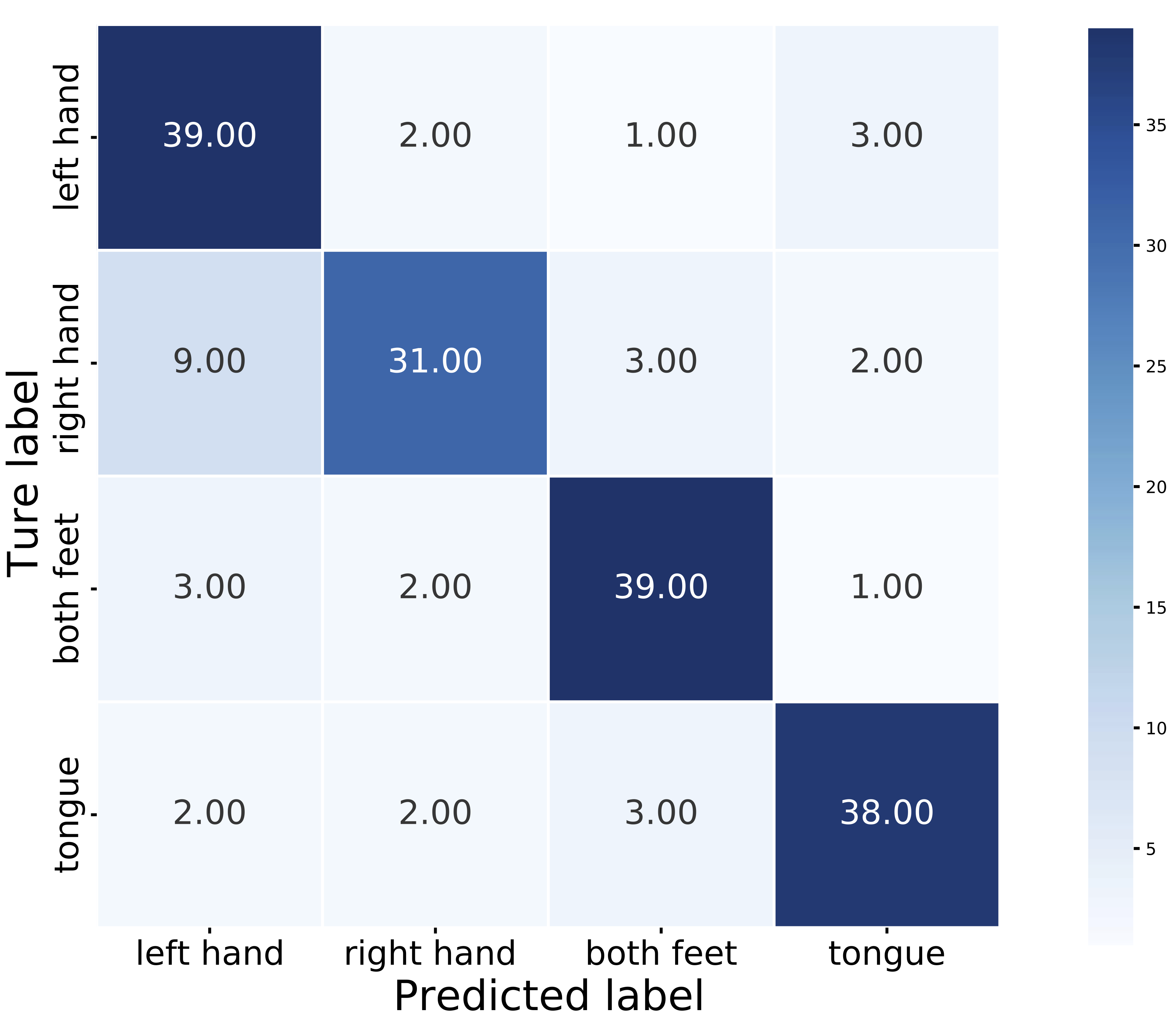}
        \caption{{\tool}}
        \label{fig:matrix-tool}
    \end{subfigure}
    \hfill
    \begin{subfigure}{0.325\textwidth}
        \centering
        \includegraphics[width=\textwidth]{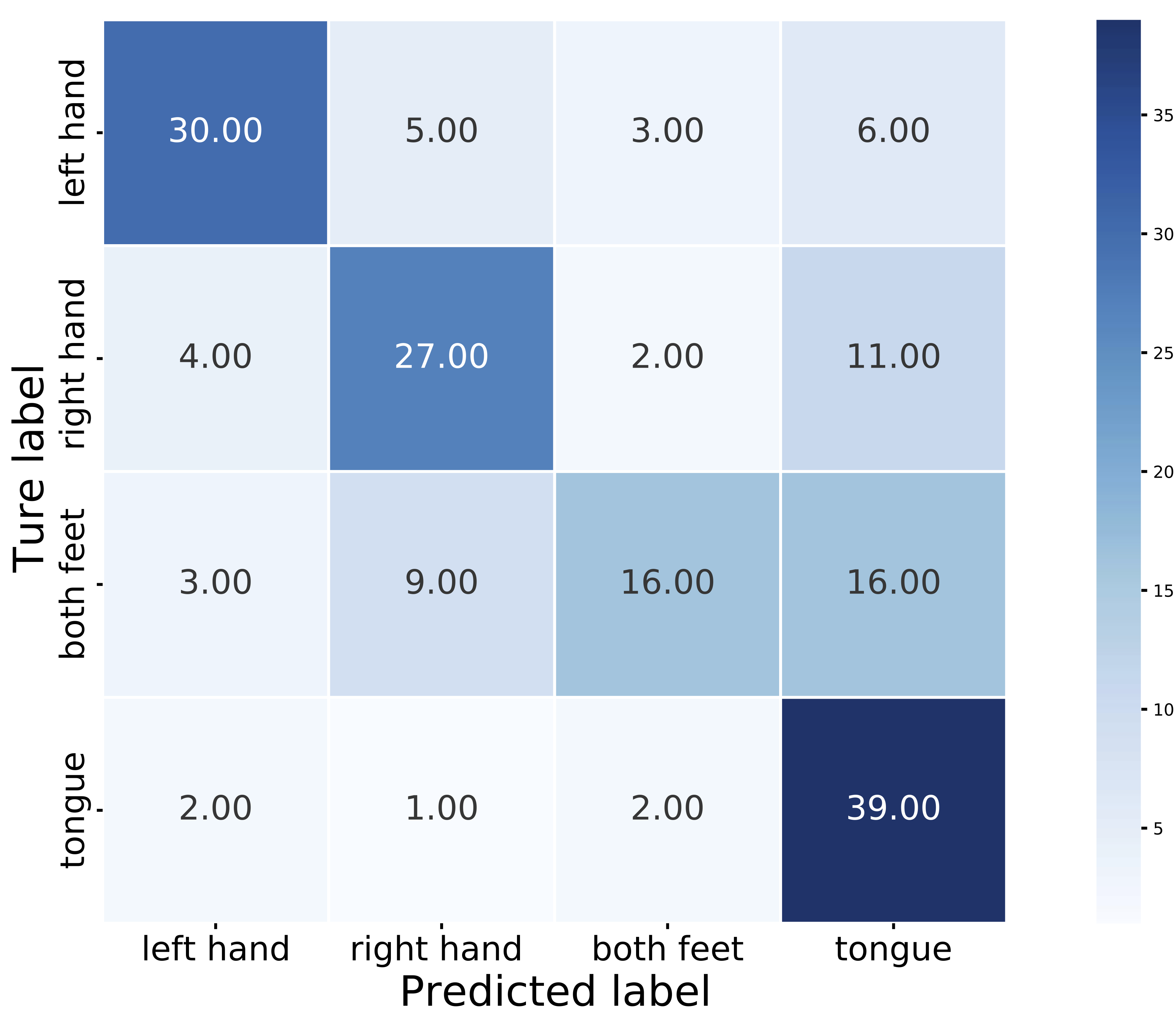}
        \caption{Tensor-CSPNet}
        \label{fig:matrix-cspnet}
    \end{subfigure}
    \hfill
    \begin{subfigure}{0.332\textwidth}
        \centering
        \includegraphics[width=\textwidth]{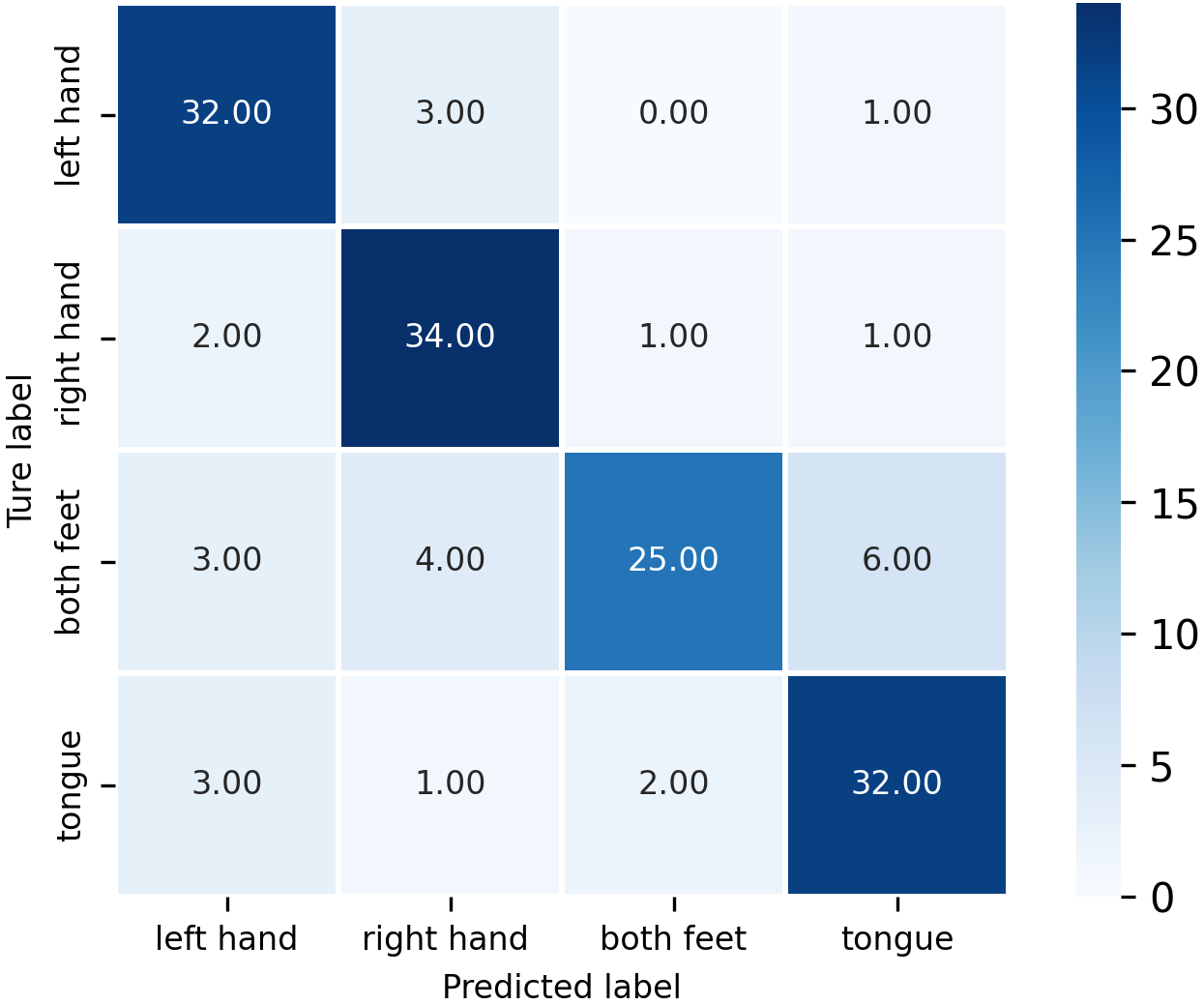}
        \caption{{\tool} on Headset-2}
        \label{fig:matrix-emotiv}
    \end{subfigure}
    
    \caption{Comparative confusion matrices illustrating MI classification performance}
    \label{fig:confusion matrix}
    
\end{figure}

To better explore the performance of both {\tool} and Tensor-CSPNet predictions, we compiled confusion matrices for a more granular view (as shown in Figure~\ref{fig:confusion matrix}). Among the set of 180 commands, {\tool} accurately predicts 147, thus achieving an accuracy rate of around 82.54\%. This significantly surpasses the accuracy demonstrated by Tensor-CSPNet and Graph-CSPNet, which achieve 62.22\% and 53.10\%, respectively.
Upon more intricate inspection,
we found that {\tool} sporadically misinterprets the Left Hand and Right Hand commands,
with nine instances of the Right Hand wrongly identified as the Left Hand
and two instances of the reverse situation.
A comparative analysis of the confusion matrices from {\tool},
Tensor-CSPNet, and Graph-CSPNet illuminates this discrepancy,
indicating that such misinterpretations occur less frequently with {\tool}.
Moreover, the commands for Both Feet
and Tongue register higher accuracy rates with {\tool},
thereby underscoring its proficiency at
translating motor-imagery recognition into
distinct control commands with promising accuracy.
This reflects the potential of {\tool} to manage real-world devices.
The confusion matrix for prototype Headset-2
is shown in Figure~\ref{fig:matrix-emotiv}.
After fine-tuning, the accuracy still reached 82\%,
which indicates that {\tool} can perform well
on different hardware devices even with different output value ranges of raw EEG data.
This compatibility and generalization demonstrate the considerable applicability of {\tool}.

Taken together, our experiments demonstrate that
{\tool} outperforms both Tensor-CSPNet and Graph-CSPNet
in terms of accuracy and parameter efficiency.
Specifically, {\tool} has $64.9K$ parameters and achieves an accuracy of 82.54\%,
while Tensor-CSPNet and Graph-CSPNet have $183.7K$ and $151.2K$ parameters and accuracies of 62.22\% and 53.10\%, respectively.
This highlights the efficiency and effectiveness of {\tool} for motor-imagery recognition tasks.

\begin{figure*}[!tb]
\centering
\includegraphics[width=0.8\textwidth]{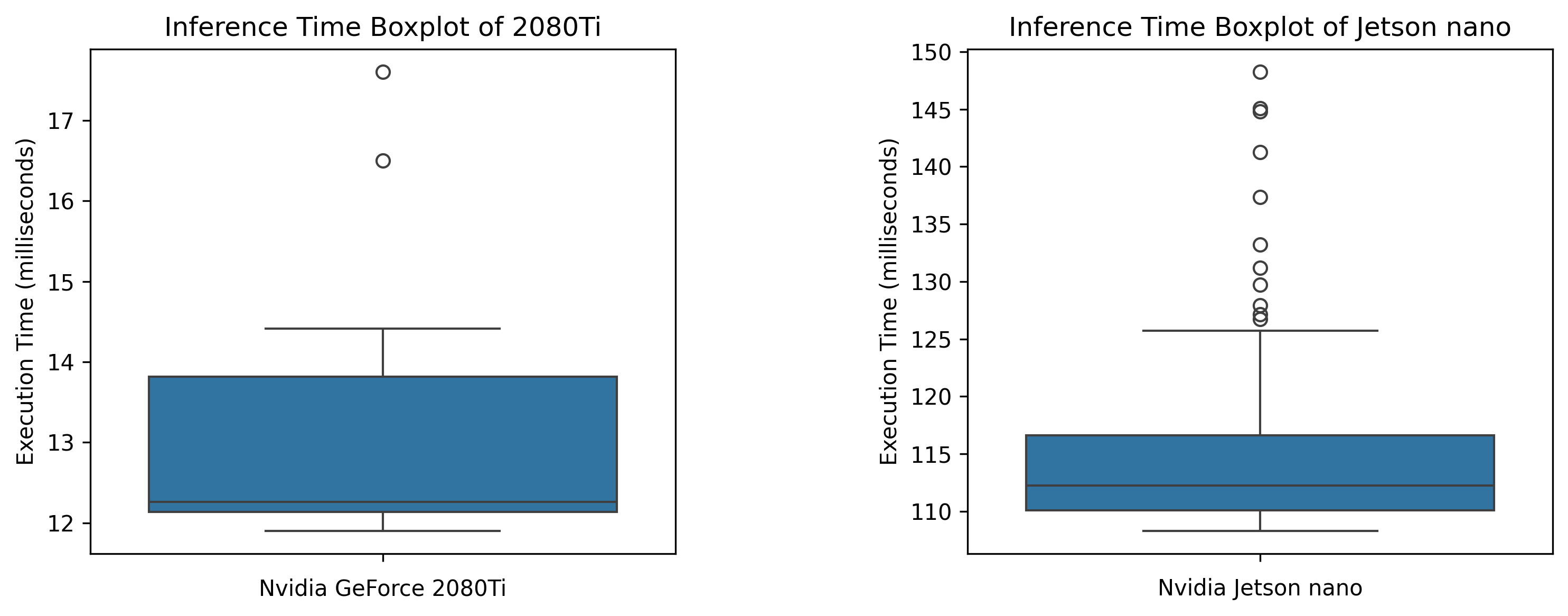}
\caption{Comparison of inference speeds on mobile and desktop devices of {\tool} }
\label{fig:mobile_infer}
\vspace{-5pt}
\end{figure*}

To further examine the application in real-world scenarios,
we deployed {\tool} on an NVIDIA Jetson Nano Development Kit (B01) for inference speed testing. The comparison of inference speeds is shown in Figure~\ref{fig:mobile_infer}.
The average inference time on this IoT edge device is 115 ms,
without code-level nor runtime-level optimization.
Comparing with a desktop computer with an NVIDIA GeForce RTX 2080Ti GPU,
which has 4352 CUDA cores, 11 GB dedicated graphic memory,
1350 MHz clock speed, and 250 Watts electrical power requirement,
Jetson Nano has 128 CUDA cores, 4 GB shared memory, and 921.6 MHz GPU clock speed,
with a 10 Watts electrical power requirement.
Though the inference speed on Jetson nano is 10$\times$ slower than on the 2080Ti,
it is still within the acceptable range,
even though the Jetson nano is small and consumes much less
power (25$\times$ less than the 2080Ti). This guarantees the feasibility of our {\tool} prototype in a real-world deployment.

\subsection{Discussion}
{\tool} outperforms Tensor-CSPNet and Graph-CSPCSPNet, due to its robust handling of lower sampling rates, selective frequency band utilization, and advanced feature extraction techniques. Below are the detailed reasons for its advantages:

\begin{enumerate}
\item Sampling Rate and Performance Impact:
Our self-collected data is sampled at a rate of 128 Hz, 
resulting in fewer sample points within a given time interval compared to public datasets sampled at 1000 Hz and 250 Hz. 
Tensor-CSPNet generally performs well when a sample contains a sufficient number of data points. 
As shown in Table~\ref{tab:SOTA Comparison}, 
Tensor-CSPNet achieves the best performance among all baselines on the MI-KU dataset, 
which is sampled at 1000 Hz. 
However, when the sampling rate decreases to 250 Hz in the BCIC-IV-2a dataset, 
it is surpassed by both Graph-CSPNet and our {\tool} model.
Graph-CSPNet excels on datasets such as MI-KU and BCIC-IV-2a, 
which have high sampling rates of 1000 Hz and 250 Hz, respectively. 
These elevated sampling rates result in the generation of 10,000 and 2500 data points 
over a 10-second interval. T
he abundance of data points is crucial for constructing detailed and effective time-frequency graphs. 
With more data points, the time-frequency representations become more granular and precise, 
allowing Graph-CSPNet to capture intricate temporal and spectral patterns within the EEG signals. 
This high level of detail is essential for the model's performance, 
enabling the identification of subtle features and variations critical for accurate classification and analysis.

Due to budget constraints, our equipment samples at only 128 Hz, 
producing only 1280 data points in the same interval.
This is just 12.8\% of the data points in the MI-KU dataset and 51.2\% of those in the BCIC-IV-2a dataset.
Tensor-CSPNet and Graph-CSPNet both suffer from reduced performance at lower sampling rates, 
due to their reliance on high-resolution data. When sampling rates decrease,
the models lose the temporal and spectral resolution necessary for capturing fine-grained features in EEG signals.
Tensor-CSPNet struggles because its feature extraction methods cannot effectively process the limited data points,
leading to a significant drop in accuracy.
Similarly, Graph-CSPNet's performance declines as the sparse data results in inadequate time-frequency graphs,
hindering the model's ability to identify complex patterns.
Our analysis does not imply that existing SOTA methods inherently underperform on data with a lower sampling rate,
but rather highlights the versatility and effectiveness of our framework in such conditions.
While sub-sampling public datasets was suggested, we believe that this could unfairly disadvantage methods
like graph-based approaches that rely on higher sampling rates to model fine-grained temporal dependencies,
making the comparisons less meaningful.
By contrast, our {\tool} model shows significant advantages, achieving an accuracy of 82.54\%.
Unlike Tensor-CSPNet and Graph-CSPNet, {\tool} is better suited for datasets with lower sampling rates,
making it more effective with our self-collected data.
This adaptability allows {\tool} to perform more robustly in our experimental setup,
demonstrating higher accuracy and reliability.
The primary reason for our model's robust performance at lower sampling rates is its design,
which efficiently captures and processes essential features from EEG signals even with fewer data points.
Our model leverages geometric deep learning and the SPD manifold,
which are adept at managing spatial correlations and extracting meaningful patterns from data.
Additionally, our efficient channel selection module and multi-head bilinear transformation strategy
contribute to the model's ability to maintain high performance in resource-constrained environments.
These components enable our model to utilize the available data effectively,
ensuring that critical information is preserved and accurately interpreted,
thus maintaining reliability and accuracy despite the lower sampling rate.

\item Selective Frequency Band Utilization: Unlike Tensor-CSPNet and Graph-CSP,
our model selectively targets the most significant frequency bands for specific BCI tasks, rather than utilizing the entire frequency spectrum.
This focused approach is advantageous because not all frequency bands contribute equally
to signal quality and task performance; some may introduce noise and reduce the model's effectiveness.
By concentrating on critical bands, our model mitigates these issues,
enhancing its accuracy and signal interpretation. This is particularly beneficial in datasets with lower sampling rates or environmental noise.

\item Enhanced Feature Extraction with a Multi-Head Approach:
The resilience of our model in datasets with fewer sample points is further
supported by an innovative feature extraction strategy that incorporates a multi-head mechanism,
similar to those found in advanced neural network architectures like transformers.
This strategy employs multiple parallel attention mechanisms,
each analyzing different facets of the input data.
This enhances the model's ability to capture a broader range of signal characteristics and ensures robust performance,
even when traditional extraction methods might falter.
This approach is particularly effective for complex BCI tasks where a nuanced interpretation of brain signals is critical.
\end{enumerate}

\section{Limitations and Future Directions}
\label{sec:future}

As a prototype, our proposed {\tool}  has limitations, which we discuss below, along with potential future directions.

\subsection{Latency Induced by Windowing Size}

As mentioned in Section~\ref{sec:module-1}, {\tool} leverages window slicing in signal processing.
Though the windowing length could be adjusted, 
it must still be an adequate length for manifold construction.
In other words, there will be an unavoidable latency between the start of the imagery of the participants and the output of the prototype prediction result.
This latency could be mitigated in several ways.
Advanced signal processing techniques such as adaptive windowing methods
could help to adjust window size dynamically for faster response times.
Meanwhile, improving our algorithm in {\tool} could help make
accurate predictions with shorter window lengths.
In future studies, we will implement these methodologies to
reduce the latency and enhance the responsiveness and efficiency of {\tool}.

\subsection{Reliability for Real-World Deployment}

This prototype is still not yet ready for real-world deployment.
As mentioned in Section~\ref{sec:rq2}, the accuracy is about 83\%.
Though it outperforms existing state-of-the-art methods,
it is still behind an acceptable rate for reliable real-world deployment.
Meanwhile, the previously mentioned latency could lead to decreased user experience,
as users will experience counter-intuitive delays in control.
Failing to output the result on time could cause serious or even fatal harm,
considering that the users might be unable to stop in time to avoid danger.
To improve the reliability of {\tool},
it is essential to explore methods for enhancing the inference time of geometric deep learning.
One approach could be to replace SPD matrix multiplication with matrix addition,
similar to the method used by \cite{zhu2024scalable}, to reduce latency on embedded devices.
In the future, we will further validate and optimize this approach in our prototype.

\subsection{Limited Encoding Spaces for Further Expansion}

Our prototype currently supports four kinds of recognition: Left Hand, Right Hand, Both Feet, and Tongue (Section~\ref{sec:rq2}). This is sufficient for encoding the four basic movements of wheelchairs,
i.e.
Forward, Stop, Left Rotation, and Right Rotation (Section~\ref{sec:eeg-wheelchair}).
While these four movements are useful, as demonstrated in our experiment,
the encoding method offers additional possibilities. 
We believe that the encoding spaces can be further expanded.
In the next stage, we will collaborate with neurologists to reveal 
the potential for encoding and designing better imagery tasks that could allow for more encoding spaces. 
We will also conduct experiments for more controlling commands such as turning remote devices on and off.

\section{Conclusion}
\label{sec:conclusion}

Brain--computer interfaces (BCIs) make it possible to bridge 
the gap between human cognition and the direct control of external devices.
In this research, we presented {\tool}, 
a prototype that leverages geometric deep learning
for EEG data interpretation through non-Euclidean metrics and the symmetric positive-definite manifold.
Our {\tool} for motor-imagery tasks
comes with a reduced model size and swift inference speed without compromising accuracy,
making it possible to run on resource-limited mobile devices.
We comprehensively validated our approach with experimental studies, including real-world scenarios,
to explore and measure the potential and abilities of {\tool}.
This investigation illuminated the transformative potential of deploying geometric deep learning in motor-imagery BCIs.
It is worth noting that gathering genuine motor-imagery signals is complex 
and requires special devices (not just scalp-contact EEG devices). 
Such devices are not currently available on the market.
We will continue to collaborate with EEG device suppliers,
and we will explore possible applications for efficient feature extraction for these kind of signals,
once such devices become available.
Meanwhile, our study also revealed the potential for applying geometric deep learning
in other sensors with non-Euclidean data, such as electrocardiograms and electromyography.
We plan to explore the implementation of geometric deep learning in more areas for new challenges and better solutions.

\begin{acks}
We are grateful for the software subscription waiver and support from Emotiv\footnote{\url{https://www.emotiv.com/}} for providing access to their EEG devices.
\end{acks}

\bibliographystyle{ACM-Reference-Format}
\bibliography{ref}

\appendix

\section{Appendix}
\label{appendix}

\subsection{\textbf{Objective Function for EEG Channel Selection Module}}
\begin{theorem}
If we let $A = (-\frac{1}{2}(G^{2}-D^{2}))$,
then $HAH$ results in a positive semi-definite (p.s.d.) matrix,
with $H=I_{m}-m^{-1}11^{T}$ representing the centering matrix.
Here, $1=[1, 1, \dots, 1]^{T} \in \mathbb{R}^{m}$,
and $I_{m}$ is an $m \times m$ identity matrix.
\label{appendix.th:Positive semi-definite property}
\end{theorem}

Theorem~\ref{appendix.th:Positive semi-definite property} is derived
from the p.s.d. property~\cite{bartlett1947multivariate}, which
states the following:
\begin{framed}
Let $A = (-\frac{1}{2}z_{ij}^{2})$,
where $z_{ij}^{2}$ represents the Euclidean distance between $i$ and $j$. Then, $HAH$ is p.s.d.
\end{framed}

For $\forall{g_{ij}}{\in}G$ and $\forall{d_{ij}}{\in}D$, $(g_{ij} - d_{ij})^{2} = g_{ij}^{2} + d_{ij}^{2} - 2g_{ij}d_{ij}$.
Given that $g_{ij}$ is expected to have a value very close to $d_{ij}$,
the expression can be approximated as $g_{ij}^{2} + d_{ij}^{2} - 2g_{ij}d_{ij} \approx g_{ij}^{2} + d_{ij}^{2} - 2d_{ij}^{2} = g_{ij}^{2} - d_{ij}^{2}$.
Thus, $(G-D)^{2} \approx (G^{2}-D^{2})$. As $(G-D)^{2}$ represents the Euclidean distance between $G$ and $D$,
we let $z_{ij}^{2} {\in}(G^{2}-D^{2})$.
Based on Theorem~\ref{appendix.th:Positive semi-definite property},
we can center the objective function:
\begin{equation}
\begin{aligned}
\hat{W} &= \mathop{\arg\min}\limits_{W} \|H(-\frac{1}{2}(G^{2}-D^{2}))H\|^{2}_{F} \\
        &= \mathop{\arg\min}\limits_{W} \|H(-\frac{1}{2}G^{2})H-H(-\frac{1}{2}D^{2})H\|^{2}_{F} \\
        &= \mathop{\arg\min}\limits_{W} \|(-\frac{1}{2}HG^{2}H)-(-\frac{1}{2}HD^{2}H)\|^{2}_{F} \\
        &=  \mathop{\arg\min}\limits_{W} \|\gamma_{G}-\gamma_{D}\|^{2}_{F}.
\label{appendix.eq.T1_W}
\end{aligned}
\end{equation}
Here, we let $\gamma_{D}=-\frac{1}{2}H D^{2}H$ and $\gamma_{G}=-\frac{1}{2}H G^{2}H$,
which are both p.s.d. and centered inner-product matrices.

\subsubsection{\textbf{Upper Bound for the Objective Function}}

\begin{small}
\begin{equation}
\begin{aligned}
\|\gamma_{G}-\gamma_{D}\|^{2}_{F} &= \sqrt{(\sum_{i=1}^{N}\sum_{j=1}^{N}|{\gamma_{G}}_{ij}-{\gamma_{D}}_{ij}|^{2})^{2}}\\
&= \sum_{i=1}^{N}\sum_{j=1}^{N}|{\gamma_{G}}_{ij}-{\gamma_{D}}_{ij}|^{2}\\
&=  \sum_{i=1}^{N}\sum_{j=1}^{N}{\gamma_{G}}_{ij}^{2} +
\sum_{i=1}^{N}\sum_{j=1}^{N}{\gamma_{D}}_{ij}^{2} -
\sum_{i=1}^{N}\sum_{j=1}^{N}{2\gamma_{G}}_{ij}2{\gamma_{D}}_{ij}\\
&= -2\left \langle{\gamma_{G}, \gamma_{D}}\right \rangle_{F} +  \|\gamma_{G}\|^{2}_{F} + \|\gamma_{D}\|^{2}_{F}
\end{aligned}
\label{eq:Frobenius decomposition}
\end{equation}
\end{small}

Based on Eq.~(\ref{eq:Frobenius decomposition}), $\hat{W}$ of Eq~(\ref{appendix.eq.T1_W}) can be further rewritten as
\begin{equation}
\begin{aligned}
\hat{W} &= \mathop{\arg\min}\limits_{W}[-2\left \langle{\gamma_{G}, \gamma_{D}}\right \rangle_{F} +  \|\gamma_{G}\|^{2}_{F} + \|\gamma_{D}\|^{2}_{F}] \\
        &= \mathop{\arg\min}\limits_{W}[-2tr({\gamma_{G}\gamma_{D}}^{T})] +  \|\gamma_{G}\|^{2}_{F} + \|\gamma_{D}\|^{2}_{F}].
\end{aligned}
\end{equation}
Since $d_{ij} = \|W^{T}(\log(X{i}) - \log(X_{j}))W\|_{F}$ ($Eq. (4)$ in the main paper) and we let $k_{ij} = \|X_{i}-X_{j}\|_{F}$, it follows that $d_{ij} \leq  k_{ij}$.
Also, $\gamma_{D} = -\frac{1}{2}HD^{2}H$ (Eq.~(\ref{appendix.eq.T1_W})) and $\gamma_{K} = -\frac{1}{2}HK^{2}H$. Thus, $ \|\gamma_{D}\|^{2}_{F} \leq  \|\gamma_{K}\|^{2}_{F}$.
We let
\begin{equation}
\hat{W}^{*} = \mathop{\arg\min}\limits_{W}[-2tr({\gamma_{G}\gamma_{D}}^{T})] +  \|\gamma_{G}\|^{2}_{F} + \|\gamma_{K}\|^{2}_{F}.
\label{eq:upper boundary}
\end{equation}
Hence, $\hat{W}$ satisfies a specific inequality:
$\hat{W} \leq \hat{W}^{*}$.

This inequality gives us an upper boundary for the objective function. Now we can replace the original objective function $\hat{W}$ with the upper boundary $\hat{W}^{*}$, allowing the term $\|\gamma_{D}\|^{2}_{F}$ associated with $W$ to be replaced by the unrelated term $\|\gamma_{K}\|^{2}_{F}$, thereby further simplifying the objective function.
~\\

\subsubsection{\textbf{Objective Function}}
In Eq.~(\ref{eq:upper boundary}),
since
$\|\gamma_{G}\|^{2}_{F}$ and $\|\gamma_{K}\|^{2}_{F}$
are unrelated to $W$, $\hat{W}$ can be rewritten as
\begin{equation}
\begin{aligned}
\hat{W} &= \mathop{\arg\min}\limits_{W}(-2tr({\gamma_{G}\gamma_{D}}^{T})) \\
& =  \mathop{\arg\min}\limits_{W}(-2\sum_{i=1}^{N}\sum_{j=1}^{N}(\gamma_{D})_{ij}(\gamma_{G})_{ij}).
\end{aligned}
\label{appendix.eq:W_1}
\end{equation}
Let $\theta_{i} = Wlog(X{i})W$ and $\theta_{j} = W\log(X_{j}))W$. Similar to Eq.~(\ref{eq:Frobenius decomposition}), we have
\begin{equation}
\begin{aligned}
\|\theta_{i}-\theta{j}\|_{F}^{2} &= \|\theta_{i}\|_{F}^{2} + \|\theta_{j}\|_{F}^{2} - 2{\left \langle{\theta_{i}, \theta_{j}}\right \rangle}_{F} \\
&= \|\theta_{i}\|_{F}^{2} + \|\theta_{j}\|_{F}^{2} - 2{(\gamma_{D})_{ij}}.
\end{aligned}
\end{equation}
Then,
\begin{equation}
2{(\gamma_{D})_{ij}} = \|\theta_{i}\|_{F}^{2} + \|\theta_{j}\|_{F}^{2} - \|\theta_{i}-\theta{j}\|_{F}^{2}.
\label{appendix.eq:2gamma}
\end{equation}
Substituting
Eq.~(\ref{appendix.eq:2gamma}) into
Eq.~(\ref{appendix.eq:W_1}), we have
\begin{small}
\begin{equation}
\begin{aligned}
\hat{W} &= \mathop{\arg\min}\limits_{W}(\sum_{i=1}^{N}\sum_{j=1}^{N}-( \|\theta_{i}\|_{F}^{2} + \|\theta_{j}\|_{F}^{2} - \|\theta_{i}-\theta{j}\|_{F}^{2})(\gamma_{G})_{ij}) \\
&= \mathop{\arg\min}\limits_{W}(\sum_{i=1}^{N}\sum_{j=1}^{N}((- \|\theta_{i}\|_{F}^{2})(\gamma_{G})_{ij}) +\sum_{i=1}^{N}\sum_{j=1}^{N}(- \|\theta_{j}\|_{F}^{2})(\gamma_{G})_{ij}) \\  & +\sum_{i=1}^{N}\sum_{j=1}^{N}(\|\theta_{i}-\theta{j}\|_{F}^{2}(\gamma_{G})_{ij}) \\
&= \mathop{\arg\min}\limits_{W}(\sum_{i=1}^{N}((- \|\theta_{i}\|_{F}^{2})\sum_{j=1}^{N}(\gamma_{G})_{ij}) +\sum_{j=1}^{N}(- \|\theta_{j}\|_{F}^{2})\sum_{i=1}^{N}(\gamma_{G})_{ij}) \\  & +\sum_{i=1}^{N}\sum_{j=1}^{N}(\|\theta_{i}-\theta{j}\|_{F}^{2}(\gamma_{G})_{ij}) \\
&= \mathop{\arg\min}\limits_{W}\sum_{i=1}^{N}\sum_{j=1}^{N}(\|\theta_{i}-\theta{j}\|_{F}^{2}(\gamma_{G})_{ij}) \\
&= \mathop{\arg\min}\limits_{W}\sum_{i=1}^{N}\sum_{j=1}^{N}(\|W^{T}log(X_{i})W-W^{T}log(X_{j})W\|_{F}^{2}(\gamma_{G})_{ij})\\
&= \mathop{\arg\min}\limits_{W}\sum_{i=1}^{N}\sum_{j=1}^{N}(\|W^{T}(log(X_{i})-log(X_{j}))W\|_{F}^{2}(\gamma_{G})_{ij})\\
&= \mathop{\arg\min}\limits_{W}\sum_{i=1}^{N}\sum_{j=1}^{N}(tr(W^{T}(log(X_{i}) - log(X_{j}))WW^{T}(log(X_{i}) - log(X_{j}))^{T}W)(\gamma_{G})_{ij}),
\end{aligned}
\label{appendix.eq:W_2}
\end{equation}
\end{small}
where the centered inner-product matrices $\sum_{i=1}^{N}(\gamma_{G})_{ij} = 0$ and $\sum_{j=1}^{N}(\gamma_{G})_{ij} = 0$.

The Rayleigh quotient provides a way to find the maximum (or minimum) eigenvalue by optimizing the ratio of a quadratic form~\cite{croot2005rayleigh}:
\begin{equation}
\lambda_{\text{max}} = \max_{\mathbf{x} \neq \mathbf{0}} \frac{\mathbf{x}^\top\mathbf{A}\mathbf{x}}{\mathbf{x}^\top\mathbf{x}},
\end{equation}
where $\lambda_{\text{max}}$ represents the largest eigenvalue of matrix A, 
and $x$ is constrained to be non-zero. 
Accordingly, in the $(t+1)^{th}$ iteration, 
the objective function for updating $\hat{W}_{t}$ can be formulated as
\begin{equation}
\begin{aligned}
    \hat{W}_{t+1} &= argmax\ tr(W^{T} \mathcal{L} W) \quad s.t. \ W^{T}W=I_{m}.
\label{eq:w_t1}
\end{aligned}
\end{equation}
Here, $\mathcal{L}$ is defined as follows:
\begin{footnotesize}
\begin{equation}
    \mathcal{L} = -\sum\limits_{i=1}^{N}\sum\limits_{j=1}^{N}\underbrace{(\gamma_{G})_{ij}}_{\mathcal{F}(G)} \underbrace{(log(X_{i}) - log(X_{j}))\hat{W}_{t}\hat{W}^{T}_{t}(log(X_{i}) - log(X_{j}))^{T}}_{\mathcal{F}(D)}.
\label{appendix.eq:object function}
\end{equation}
\end{footnotesize}

For an orthogonal matrix
$W$ and a symmetric matrix $L$, the Rayleigh quotient~\cite{croot2005rayleigh} indicates that the columns of
$W$ can be chosen as the eigenvectors of
$L$ to maximize the trace
$tr(W^{T}LW)$. Since the columns of an orthogonal matrix form an orthonormal basis, these vectors correspond to the eigenvalues of
$L$.
Mathematically, if we wish to maximize
$tr(W^{T}LW)$, we select the columns of
$W$ to be the eigenvectors of
$L$ associated with its largest eigenvalues. Then,
$tr(W^{T}LW)$ becomes a diagonal matrix with the eigenvalues of
$L$ on its diagonal. Due to the orthogonality of
$W$, the trace (the sum of the diagonal elements) is maximized.

Therefore, to obtain
$\hat{W}_{t+1}$, we just need to take the eigenvectors corresponding to the maximum $m$ eigenvalues of $\mathcal{L}$.
The expression \(\mathcal{L}\) comprises functions of \(G\) and \(D\), denoting a unique link between the geometric and Euclidean distances among EEG channels. By conducting eigenvalue decomposition on \(\mathcal{L}\), we unveil the importance of the corresponding eigenvectors in the transformation, employing the indices of the top \(d\) eigenvectors to discern the most significant \(d\) channels.

\end{document}